\documentclass{article}

\PassOptionsToPackage{numbers,sort&compress}{natbib}
\usepackage[preprint]{arxiv_preprint}

\usepackage[utf8]{inputenc}
\usepackage[T1]{fontenc}
\usepackage{hyperref}
\usepackage{url}
\usepackage{booktabs}
\usepackage{amsfonts}
\usepackage{amsmath}
\usepackage{amssymb}
\usepackage{amsthm}
\usepackage{nicefrac}
\usepackage{microtype}
\usepackage{xcolor}
\usepackage{graphicx}
\usepackage{multirow}
\usepackage{array}
\usepackage{enumitem}
\usepackage{longtable}
\usepackage{placeins}
\usepackage{float}

\newtheorem{definition}{Definition}

\title{Steerable but Not Decodable: Function Vectors\\
       Operate Beyond the Logit Lens}

\author{Mohammed Suhail B Nadaf\\
  Independent\\
  \texttt{suhailnadaf509@gmail.com}\\
}

\begin{document}

\maketitle

\begin{abstract}
Function vectors (FVs)---mean-difference directions extracted from in-context learning demonstrations---can steer language model behavior when added to the residual stream. We hypothesized that FV steering failures would coincide with information absence: the parameter-free logit lens (the model's own unembedding applied at intermediate layers) would fail wherever steering fails. We were wrong. In the largest cross-template FV transfer study to date---4{,}032 directed pairs across 12 tasks, 6 models from 3 families (Llama-3.1-8B, Gemma-2-9B, Mistral-7B-v0.3; base and instruction-tuned), and 8 templates per task---we find the \emph{opposite} dissociation: FV steering succeeds even when the logit lens cannot decode the correct answer at any intermediate layer. This \emph{steerability-without-decodability} pattern is universal: steering accuracy meets or exceeds logit-lens accuracy for every task on every model, with gaps as large as $-0.91$; the predicted converse cell (decodable without steerable) contains only 3 of 72 task$\times$model instances. FV vocabulary projection further reveals that FVs achieving $>$0.90 steering accuracy still project to incoherent token distributions, indicating FVs encode \emph{computational instructions} rather than \emph{answer directions}. The dissociation survives a tuned-lens robustness check: per-layer diagonal affine translators close only 1 of 14 steerable-not-decodable gaps (93\% persist) across model families with a $\sim$10$\times$ range in dialect magnitude. A nonlinear-decoder follow-up (a 2-layer MLP probe with a Hewitt \& Liang control task) decomposes the SAND set: 5 of 10 cases close (information was nonlinearly encoded) and 5 persist as \emph{invisible to every decoder we tried}---the dissociation does not reduce to a logit-lens-capacity artifact. FVs intervene optimally at early layers (L2--L8) while logit-lens readability emerges only at late layers (L28--L32)---a 20+-layer temporal gap. The previously reported negative cosine--transfer correlation ($r = -0.572$) dissolves at $4\times$ task scale to pooled $r \in [-0.20, +0.13]$; cosine adds $\Delta R^2 \leq 0.011$ beyond task identity. Activation patching causally confirms this difficulty hierarchy, and post-steering logit-lens analysis exposes a model-family mechanistic divergence: Mistral FVs rewrite intermediate representations (delta up to $+0.91$), while Llama and Gemma FVs produce near-zero changes despite equally successful steering. These findings challenge the assumption that additive steering operates through linearly decodable subspaces, with direct implications for safety monitoring of steering interventions.
\end{abstract}

\section{Introduction}
\label{sec:intro}

Activation steering---adding a learned direction to a language model's hidden states at inference time---is an emerging tool for controlling large language model (LLM) behavior without retraining, with applications spanning refusal induction~\citep{arditi2024refusal}, truthfulness enhancement~\citep{li2024inference}, and representation engineering~\citep{zou2023representation}. The approach rests on the \emph{linear representation hypothesis} (LRH)~\citep{park2024geometry}: task-relevant behaviors correspond to linear directions in activation space, so a function vector (FV)~\citep{hendel2023context, todd2024function} extracted from in-context learning (ICL) demonstrations should reliably steer the model when added to its hidden states.

For safety-critical deployment the relevant test is not whether an FV works on the same prompt template from which it was extracted, but whether it transfers to \emph{different} template formulations of the same task: an adversary need only rephrase a prompt to evade a template-specific intervention~\citep{sclar2023quantifying, wei2024assessing}. Yet prior FV work evaluates almost exclusively in-distribution (IID)~\citep{hendel2023context, todd2024function, liu2023context}. A preliminary study~\citep{nadaf2025preliminary} evaluated cross-template transfer across 3 tasks and 8 templates on a single 8B base model, reporting a Simpson's paradox: an aggregate negative correlation ($r = -0.572$) between cosine alignment and transfer accuracy that dissolved within individual tasks, driven by the fact that 2 of 3 tasks had near-zero IID accuracy. That study had several limitations---3 tasks (1 working, 2 failing), one model family, shared template surface forms across tasks, no diagnostic separating information absence from intervention failure, and hardcoded analysis thresholds.

We address all of these limitations: 12 tasks across 5 principled computational categories, 6 models from 3 families, task-specific templates, logit-lens analysis, FV vocabulary projection, independent baselines, and data-derived thresholds. We frame four research questions:
\begin{enumerate}[noitemsep,leftmargin=*]
\item For which computational task types does mean-difference FV extraction succeed, and for which does it fail?
\item Is cross-template transfer success predicted by geometric alignment (cosine similarity), or by IID steering efficacy?
\item When FV steering succeeds, can the model's own logit lens decode the correct answer at intermediate layers---or does the FV operate through pathways invisible to the model's output head?
\item How do these patterns vary across model families and between base and instruction-tuned variants?
\end{enumerate}

Question 3 is the most novel---and produced the most surprising answer. Combining FV steering with the logit lens at each intermediate layer, we construct a $2\times 2$ matrix: \{logit lens decodes, fails\} $\times$ \{FV steering succeeds, fails\}. We expected the upper-right cell---\emph{decodability without steerability}---to be populated. Instead, it is empty on 4 of 6 models and contains only 3 of 72 task$\times$model instances (all Mistral). The actual finding is the lower-left cell---\emph{steerability without decodability}---with 10 of 72 instances spanning all families except Gemma Base. FVs steer behavior through subspaces the model's own unembedding cannot decode at any intermediate layer. Unlike linear probing, the logit lens uses no learned parameters, sidestepping the probe-complexity critique~\citep{hewitt2019designing}.

\paragraph{Contributions.}
\begin{enumerate}[noitemsep,leftmargin=*]
\item \textbf{The largest cross-template FV transfer study to date.} 12 tasks, 6 models, 672 directed pairs per model (4{,}032 total); pre-registered difficulty predictions with honest assessment of prediction failures.
\item \textbf{Systematic quantification of the steerability-without-decodability dissociation.} Additive FV steering succeeds where the model's own unembedding cannot decode the correct answer at any layer; the predicted opposite pattern has only 3/72 instances. Closely related observations exist~\citep{todd2024function, ward2025reasoning, sinii2025steering, arad2025saes, makelov2024subspace, she2025linear}; we organize them into a single $2 \times 2$ matrix and quantify the gap with a tuned-lens robustness check.
\item \textbf{Tuned-lens validation.} Per-layer diagonal affine translators close only 1 of 14 steerable-not-decodable gaps (93\% persist), ruling out representational dialect across families with a $\sim$10$\times$ correction-magnitude range.
\item \textbf{MLP-probe decoder ladder.} A 2-layer MLP probe with Hewitt \& Liang control selectivity decomposes the 10 logit-lens SAND cells into 5 closed by nonlinear decoding and 5 invisible to every decoder we tried; the Hewitt \& Liang gate vetoes 6 false positives on \texttt{object\_color} where probes attain $>$0.82 top-10 by exploiting label-marginal shortcuts.
\item \textbf{Direct evidence that FVs encode computational instructions, not answer directions.} FV vocabulary projections are universally incoherent even at $>$0.90 steering accuracy.
\item \textbf{The reported negative cosine--transfer correlation was a small-sample artifact.} At $4\times$ task scale the pooled $r$ dissolves to $|r| \leq 0.20$; cosine adds $\Delta R^2 \leq 0.011$ beyond task identity.
\item \textbf{A novel model-family mechanistic divergence.} Post-steering logit-lens analysis shows Mistral FVs rewrite intermediate representations (delta up to $+0.91$), while Llama and Gemma FVs produce near-zero changes on steerable-not-decodable tasks despite equally successful steering---evidence for two distinct mechanisms.
\item \textbf{Activation patching localizes cross-template transfer.} Tasks with strong IID steering achieve perfect recovery (1.000) at specific residual-stream layers; genuinely hard tasks show zero recovery everywhere, confirming that the difficulty hierarchy reflects causal localization rather than statistical correlation.
\end{enumerate}

\FloatBarrier
\section{Background and Related Work}
\label{sec:related}

\paragraph{Function vectors and in-context learning.}
\citet{hendel2023context} demonstrated that ICL creates task-specific directions in activation space; \citet{todd2024function} extended this to function vectors identified via causal mediation over attention heads and observed (their Section~3.2) that FVs decoded through the unembedding produce token distributions ``not enough on their own to construct a working function vector.'' \citet{liu2023context} proposed in-context vectors for more controllable steering. \citet{dasilva2025steering} showed at scale (36 models) that FV reliability varies substantially across families. We extend this work to a fully bidirectional cross-template matrix (56 directed pairs per task) across 12 tasks and 6 models, generalize the vocabulary-projection observation across families, and tie it to a tuned-lens robustness check.

\paragraph{Activation steering and the linear representation hypothesis.}
\citet{turner2023activation} introduced activation addition as an optimization-free steering method. \citet{panickssery2023steering} and \citet{rimsky2024steering} formalized contrastive activation addition (CAA), the mean-of-differences approach we adopt; \citet{im2025unified} compared CAA, PCA, and classifier-based extraction and found mean-of-differences theoretically and empirically strong. \citet{zou2023representation} proposed representation engineering as a general control framework; \citet{arditi2024refusal} showed refusal is mediated by a single direction; \citet{marks2023geometry, tigges2023linear, hernandez2024linearity} found linear structure in truth-value, sentiment, and relation representations. \citet{park2023linear, park2024geometry} articulated the LRH, distinguishing an \emph{unembedding representation} (probing/decodability) from an \emph{embedding representation} (additive steering) and unifying them via a causal inner product. The reliability of additive steering has since been scrutinized: \citet{tan2024analysing} found steerability is largely dataset-level with substantial per-sample anti-steerability, \citet{braun2025understanding} showed prompt-type variants yield directionally distinct yet behaviorally comparable vectors, and \citet{goldmanwetzler2024orthogonal} found $>$800 nearly orthogonal vectors all inducing the same behavior---geometric similarity is not a reliable behavioral proxy. Building on the unembedding/embedding distinction, we empirically decompose the LRH into \emph{linear decodability} and \emph{linear steerability} and show these come apart in the opposite direction from naive intuition.

\paragraph{Probing, the logit lens, and vocabulary projection.}
\citet{belinkov2022probing} surveyed probing classifiers; \citet{hewitt2019designing} raised the probe-complexity concern. The logit lens~\citep{nostalgebraist2020logitlens} sidesteps this by projecting intermediate activations through the model's own final LayerNorm and unembedding matrix $W_U$, requiring no learned parameters. It is well-known to fail at early layers~\citep{belrose2023eliciting}, motivating the tuned lens with per-layer affine corrections, which we apply as a conservative diagonal robustness check (\S\ref{sec:tunedlens}). Closely related observations exist: \citet{ward2025reasoning} noted base-model steering vectors ``never map to backtracking logits at early-to-mid layers''; \citet{arad2025saes} found at the SAE level that input (decodability) and output (steerability) features ``rarely co-occur''; \citet{she2025linear} reported a temporal version of the dissociation during training (decodability before steerability); \citet{makelov2024subspace} showed steering can operate via ``dormant pathways'' disconnected from where information appears to be encoded. Our contribution is a systematic, multi-model, multi-task quantification of this gap with a tuned-lens robustness check.

\paragraph{Mechanistic interpretability and prompt sensitivity.}
Our patching methodology follows causal tracing~\citep{meng2022locating, conmy2023towards}. \citet{wang2022interpretability} analyzed indirect-object identification circuits, \citet{olsson2022context} identified induction heads, and \citet{geva2023dissecting} dissected factual recall. \citet{sclar2023quantifying} quantified prompt-formatting sensitivity at the behavioral level; we extend prompt sensitivity analysis to the internal geometry of steering vectors. Simpson's paradox~\citep{simpson1951interpretation} in steering evaluation was first identified in the preliminary study above; we test whether it persists at $4\times$ scale.

\FloatBarrier
\section{Experimental Setup}
\label{sec:setup}

\subsection{Task Battery}
We design 12 tasks spanning 5 computational categories (Table~\ref{tab:tasks}). The taxonomy is motivated by the hypothesis that different computation types engage different transformer circuits~\citep{geva2023dissecting, todd2024function} and may exhibit different FV geometry and transfer behavior. Categories correspond to qualitatively different computational demands: \textbf{Lexical Retrieval} (stored word-level associations), \textbf{Factual Retrieval} (world-knowledge facts), \textbf{Morphological Transform} (regular grammatical rules), \textbf{Character/Surface} (character-level manipulation; \emph{predicted to fail}), and \textbf{Compositional/Semantic} (multi-token semantic rewriting; predicted to fail).

\begin{table}[!htbp]
\caption{Task battery: 12 tasks across 5 computational categories. ``Expected IID'' is the pre-registered difficulty prediction (Appendix~\ref{app:predictions}), formulated before running any experiments. $n$ is the number of input--output pairs. Eval: substring match (case-sensitive for \texttt{capitalize}); polarity classification for \texttt{sentiment\_flip} (\S\ref{sec:logit_lens}, Appendix~\ref{app:stat}). Tok: maximum new tokens.}
\label{tab:tasks}
\centering
\small
\begin{tabular}{@{}llrlllr@{}}
\toprule
\textbf{Category} & \textbf{Task} & $n$ & \textbf{Eval} & \textbf{Expected IID} & \textbf{Role} & \textbf{Tok} \\
\midrule
\multirow{3}{*}{Lexical Retrieval} & \texttt{antonym} & 95 & substr & 0.45--0.60 & Positive control & 5 \\
& \texttt{synonym} & 88 & substr & 0.20--0.40 & Moderate test & 5 \\
& \texttt{hypernym} & 86 & substr & 0.25--0.45 & Moderate test & 5 \\
\midrule
\multirow{3}{*}{Factual Retrieval} & \texttt{country\_capital} & 90 & substr & 0.40--0.65 & Positive control & 5 \\
& \texttt{english\_spanish} & 88 & substr & 0.25--0.50 & Moderate test & 5 \\
& \texttt{object\_color} & 85 & substr & 0.30--0.55 & Moderate test & 5 \\
\midrule
\multirow{2}{*}{Morphological} & \texttt{past\_tense} & 90 & substr & 0.35--0.55 & Positive control & 5 \\
& \texttt{plural} & 90 & substr & 0.30--0.50 & Moderate test & 5 \\
\midrule
\multirow{3}{*}{Character / Surface} & \texttt{capitalize} & 84 & case-s. & 0.00--0.05 & Negative control & 5 \\
& \texttt{first\_letter} & 86 & substr & 0.05--0.20 & Boundary test & 3 \\
& \texttt{reverse\_word} & 80 & substr & 0.00--0.08 & Negative control & 5 \\
\midrule
Compositional & \texttt{sentiment\_flip} & 60 & substr & 0.00--0.05 & Negative control & 10 \\
\bottomrule
\end{tabular}
\end{table}

\subsection{Templates and Models}
For each task we design 8 task-specific templates across 4 style categories: Natural ($\times 2$), Symbolic ($\times 2$), Question ($\times 2$), and Formal ($\times 2$). Critically, no two tasks share any template string. Every template contains at least one task-specific keyword (``antonym,'' ``capital,'' ``past tense''), and symbolic templates use unique function names (\texttt{antonym()}, \texttt{capital()}). This eliminates the shared-template confound identified in the preliminary study. Examples by style: ``The opposite of \{X\} is'' (Natural), ``antonym(\{X\}) ='' (Symbolic), ``What is the antonym of \{X\}?'' (Question), ``Antonym relation: \{X\} maps to'' (Formal). All 96 templates are listed in Appendix~\ref{app:templates}. Eight templates yield $8 \times 7 = 56$ directed cross-template pairs per task, $12 \times 56 = 672$ pairs per model, and $4{,}032$ total. We evaluate 6 models from 3 families, size-matched at 7--9B parameters: Llama-3.1-8B (32 layers, $d_{\text{model}}{=}4096$), Gemma-2-9B (42 layers, $d_{\text{model}}{=}3584$), and Mistral-7B-v0.3 (32 layers, $d_{\text{model}}{=}4096$), each in Base and Instruct variants.

\subsection{Function Vector Extraction and Steering}
We extract FVs via the mean-of-differences method---the standard formulation of contrastive activation addition~\citep{panickssery2023steering, rimsky2024steering, im2025unified}, applied in the function-vector setting of \citet{todd2024function}. For task $t$, template $k$, and layer $\ell$:
\begin{equation}
\mathrm{FV}_{t,k,\ell} \;=\; \mathbb{E}\bigl[\mathbf{h}^{(\text{pos})}_\ell\bigr] \;-\; \mathbb{E}\bigl[\mathbf{h}^{(\text{neg})}_\ell\bigr],
\label{eq:fv}
\end{equation}
where $\mathbf{h}^{(\text{pos})}_\ell$ and $\mathbf{h}^{(\text{neg})}_\ell$ are residual-stream activations at layer $\ell$ and the final-token position for positive (15 correct ICL demonstrations) and negative (15 zero-shot prompts) inputs. At inference we add the FV with scalar multiplier $\alpha$ to the final-token-position residual stream:
\begin{equation}
\mathbf{h}'_\ell \;=\; \mathbf{h}_\ell \;+\; \alpha \cdot \mathrm{FV}_{t,k,\ell}.
\label{eq:steer}
\end{equation}
We sweep $\alpha \in \{0.5, 1.0, 1.5, 2.0, 2.5, 3.0, 4.0, 5.0\}$ with adaptive refinement around the best $\alpha$ and report best accuracy across all $\alpha$ and layers. \textbf{IID} evaluation uses the same template for extraction and test; \textbf{OOD} (cross-template) uses different templates (56 directed pairs per task).

\begin{definition}[IID gating]
\label{def:iid_gate}
A task $t$ on model $m$ \emph{passes the IID gate} if its mean IID accuracy across templates exceeds $\tau = 0.10$. Tasks failing the gate are excluded from causal analyses (activation patching), since for these tasks FV extraction itself has not produced a working steering direction.
\end{definition}

\subsection{Logit Lens and FV Vocabulary Projection}
\label{sec:logit_lens}
To distinguish \emph{information absence} from \emph{intervention failure} we apply two parameter-free analyses to zero-shot activations.

\paragraph{Logit lens.} For each task, template, and layer $\ell$ we project the zero-shot residual stream through the model's own final LayerNorm and unembedding:
\begin{equation}
\hat{\mathbf{y}}_\ell \;=\; \mathrm{LayerNorm}_{\text{final}}(\mathbf{h}_\ell) \cdot W_U + \mathbf{b}_U.
\label{eq:logit_lens}
\end{equation}
We measure top-$k$ accuracy ($k \in \{1, 5, 10\}$): the fraction of examples where the correct output token appears among the $k$ highest-scoring tokens. No training data or hyperparameters are required. For \texttt{sentiment\_flip}, where first-token accuracy is uninformative (both polarities can begin with arbitrary tokens), we use a contrast-vector polarity classification (Appendix~\ref{app:stat}); we flag this explicitly in every figure and table caption that includes \texttt{sentiment\_flip} readability.

\paragraph{FV vocabulary projection.} We project each FV through the same pipeline:
\begin{equation}
\hat{\mathbf{z}}_{t,k,\ell} \;=\; \mathrm{LayerNorm}_{\text{final}}(\mathrm{FV}_{t,k,\ell}) \cdot W_U + \mathbf{b}_U.
\label{eq:fv_vocab}
\end{equation}
The top-50 tokens by score reveal what the FV ``pushes toward''; we report the \emph{correct-output fraction}: the proportion of top-50 tokens matching any correct task output.

\paragraph{Tuned-lens robustness check.}
\label{sec:tunedlens}
To rule out the possibility that the logit lens misses task-relevant information due to \emph{representational dialect}---per-layer differences in how activations are oriented relative to the unembedding frame---we apply the tuned lens~\citep{belrose2023eliciting} as a robustness check. We train per-layer diagonal affine translators $T_\ell(\mathbf{h}) = (\mathbf{1} + \boldsymbol{\delta}_\ell) \odot \mathbf{h} + \mathbf{b}_\ell$, identity-initialized and trained to minimize MSE between the translated layer-$\ell$ activation and the final-layer activation: $\|T_\ell(\mathbf{h}_\ell) - \mathbf{h}_L\|^2$. Each translator has $2 d_{\text{model}}$ parameters (${\sim}$8K for Llama and Mistral, ${\sim}$7K for Gemma) against ${\sim}$8{,}176 pooled training examples. The diagonal constraint is deliberately conservative: it captures per-feature scaling and offset corrections (the dominant source of dialect) without learned classification capacity, preserving the spirit of the parameter-free logit lens while correcting known geometric distortions.

\subsection{The Steerability--Decodability Matrix}
Together these analyses yield the $2\times 2$ matrix in Table~\ref{tab:matrix_def}: each task$\times$model instance is classified by whether FV steering accuracy and best-layer logit-lens top-10 accuracy exceed $\tau = 0.10$.

\begin{table}[!htbp]
\caption{The steerability--decodability matrix. ``Logit lens succeeds'' means best-layer top-10 accuracy exceeds 0.10; ``FV steering succeeds'' means mean IID accuracy exceeds $\tau = 0.10$. Naive intuition predicts the upper-right cell (decodable without steerable) would be populated. Our data invert the prediction.}
\label{tab:matrix_def}
\centering
\small
\begin{tabular}{@{}lll@{}}
\toprule
 & \textbf{FV steering succeeds} & \textbf{FV steering fails} \\
\midrule
\textbf{Logit lens succeeds} & Decodable + steerable & Decodable without steerability \\
\textbf{Logit lens fails} & \textbf{Steerable without decodability} & Information absent \\
\bottomrule
\end{tabular}
\end{table}

\subsection{Baselines and Geometric Analysis}
\textbf{Baselines:} zero-shot (no demonstrations, no steering) and few-shot ICL (5 demonstrations, no steering), establishing task difficulty independently of FV extraction. \textbf{Geometric analysis:} for each task at each layer we compute pairwise cosine similarity between template FVs, the alignment--transfer Pearson correlation $r$, and a hierarchical regression $\mathrm{OOD\_accuracy} \sim \mathrm{task\_identity} + \mathrm{cosine}$, where task identity is encoded as the per-task mean transfer accuracy; we report the incremental $\Delta R^2$ from adding cosine.

\FloatBarrier
\section{Results: Steering Efficacy and Cross-Template Transfer}
\label{sec:results}

\subsection{In-Distribution Steering}
\label{sec:iid}

\begin{figure}[!htbp]
\centering
\includegraphics[width=\linewidth]{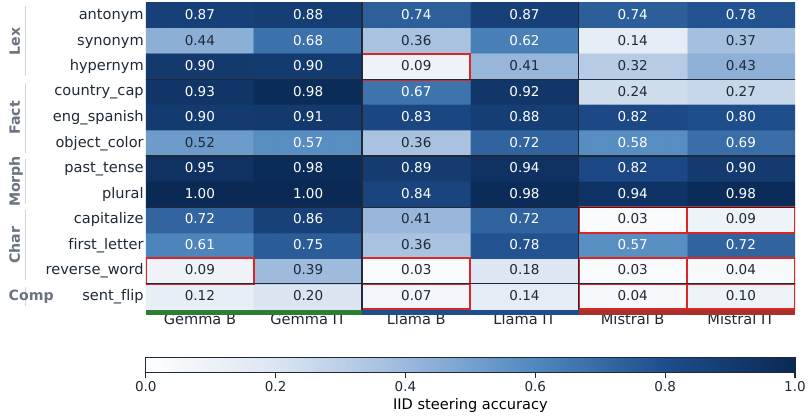}
\caption{IID steering accuracy across all 12 tasks and 6 models (mean across 8 templates at the best layer and best $\alpha$). \textbf{Red borders} mark cells failing the IID gate ($\tau = 0.10$); \textbf{orange stars} indicate falsified negative-control predictions (\texttt{capitalize}, \texttt{first\_letter}). Morphological tasks (\texttt{past\_tense}, \texttt{plural}) dramatically exceed predictions across all models. \texttt{hypernym} shows the largest cross-model divergence (0.085 on Llama Base vs.\ 0.902 on Gemma Base), indicating task category predicts FV success only as a proxy for model competence. Per-template numerical values for Llama Base are in Appendix~\ref{app:per_template}.}
\label{fig:iid_heatmap}
\end{figure}

Figure~\ref{fig:iid_heatmap} reports IID steering accuracy for all 12 tasks across 6 models; full per-family baselines are in Appendix~\ref{app:baselines}.

Morphological tasks dramatically exceeded predictions: \texttt{past\_tense} 0.823--0.982 across models (predicted 0.35--0.55), \texttt{plural} 0.845--1.000 (predicted 0.30--0.50)---the strongest positive controls in the battery. Two of three character ``negative controls'' falsified the prediction: \texttt{capitalize} reached 0.413--0.865 on Llama and Gemma (though near-zero, 0.030--0.088, on Mistral) and \texttt{first\_letter} 0.358--0.777 across all 6 models; only \texttt{reverse\_word} (0.028--0.393) behaved as predicted. These 8B+ networks may have learned character-level operations as pattern-matching over token-position embeddings rather than the sequential algorithmic procedures our taxonomy assumed.

\texttt{hypernym} shows the largest cross-model divergence: 0.085 (Llama Base) vs.\ 0.902 (Gemma Base)---a range of 0.817. Task category therefore predicts FV success only as a proxy for model competence, not as an intrinsic computational property. Zero-shot baseline accuracy correlates strongly with IID steering success ($r = 0.72$--$0.87$ across models; Appendix~\ref{app:baselines}), confirming that FVs redirect existing capabilities rather than teaching new ones. Instruction tuning consistently improves steering accuracy: mean $\Delta = +0.21$ (Llama), $+0.09$ (Gemma), $+0.07$ (Mistral); full per-task effects are in Appendix~\ref{app:instruct}.

FV steering is almost never destructive at the \emph{aggregate} level: across all 6 models, the rate at which mean steered accuracy falls below zero-shot is 0--1\% (mean improvement over zero-shot: $+$0.25 to $+$0.41). This is an aggregate-task statistic and is therefore compatible with prior reports of substantial \emph{per-sample} anti-steerability~\citep{tan2024analysing, braun2025understanding}: positive effects can dominate the mean even when individual examples are pushed in the wrong direction.

\subsection{Cross-Template Transfer}
\label{sec:transfer}

For tasks passing the IID gate, cross-template transfer barely degrades. Table~\ref{tab:transfer} reports IID vs.\ OOD accuracy on Llama-3.1-8B Base. For top-performing tasks (\texttt{plural}, \texttt{antonym}, \texttt{past\_tense}) the IID--OOD gap is consistently below 0.10, indicating genuine template-invariant steering. On Gemma models, \texttt{plural} transfers nearly perfectly (IID $= 0.998$--$1.000$, OOD $= 0.994$--$0.996$, gap $< 0.005$). The largest gap is \texttt{english\_spanish} ($-0.176$ on Llama Base), driven by a few source templates with weak FV extraction. Source-template IID quality strongly predicts OOD transfer ($r = 0.80$--$0.93$ across all 6 models): cross-template ``failure'' is primarily a source-quality problem, not a template-compatibility problem.

\begin{table}[!htbp]
\caption{Cross-template transfer for Llama-3.1-8B Base (tasks passing the IID gate). OOD accuracy is the mean across all 56 directed cross-template pairs at the best layer and $\alpha$. Per-model transfer tables are in Appendix~\ref{app:transfer}.}
\label{tab:transfer}
\centering
\small
\begin{tabular}{@{}llccc@{}}
\toprule
\textbf{Cat.} & \textbf{Task} & \textbf{IID} & \textbf{OOD} & \textbf{Gap} \\
\midrule
Lex. & \texttt{antonym} & 0.738 & 0.744 & $+$0.007 \\
Lex. & \texttt{synonym} & 0.357 & 0.316 & $-$0.041 \\
Fact. & \texttt{country\_capital} & 0.665 & 0.632 & $-$0.033 \\
Fact. & \texttt{english\_spanish} & 0.833 & 0.656 & $-$0.176 \\
Fact. & \texttt{object\_color} & 0.357 & 0.365 & $+$0.008 \\
Morph. & \texttt{past\_tense} & 0.892 & 0.793 & $-$0.099 \\
Morph. & \texttt{plural} & 0.845 & 0.887 & $+$0.042 \\
Char. & \texttt{capitalize} & 0.413 & 0.348 & $-$0.065 \\
Char. & \texttt{first\_letter} & 0.357 & 0.340 & $-$0.017 \\
\bottomrule
\end{tabular}
\end{table}

\paragraph{Within-style vs.\ across-style transfer.} Template surface form (Natural, Symbolic, Question, Formal) does not affect FV transfer. Across all 6 models, within-style and across-style mean transfer accuracies are statistically indistinguishable, $p > 0.30$ on every model (Llama Base: 0.424 vs.\ 0.438, $p = 0.72$; Gemma IT: 0.723 vs.\ 0.735, $p = 0.72$; Mistral IT: 0.456 vs.\ 0.478, $p = 0.58$). FVs capture an abstract, style-invariant computational instruction not tied to the extraction-template surface form. Full per-model statistics are in Appendix~\ref{app:style}.

\subsection{Geometric Analysis: Alignment and Transfer}
\label{sec:geometric}

\paragraph{The previously reported Simpson's paradox dissolves at 12-task scale.}
The pooled Pearson correlation between cosine similarity and OOD transfer accuracy now ranges from $r = -0.199$ (Mistral Base, $p = 1.2 \times 10^{-96}$) to $r = +0.126$ (Gemma IT, $p = 3.0 \times 10^{-51}$), with Llama Base at $r = +0.013$ ($p = 0.16$, ns). Within-task correlations are scattered around zero (Llama Base mean within-task $r = +0.007$, range $[-0.51, +0.28]$). Hierarchical regression confirms the dissolution (Table~\ref{tab:hierarchical}): task identity alone explains 40--58\% of variance in transfer accuracy ($R^2 = 0.402$ on Llama Base; 0.585 on Mistral IT), and cosine adds at most $\Delta R^2 = 0.011$---less than 1.1 percentage points.

\begin{table}[!htbp]
\caption{Hierarchical regression: task identity vs.\ cosine as predictors of OOD transfer accuracy (computed from raw cross-template pairs). Cosine contributes at most 1.1 percentage points of $R^2$ beyond task identity on every model.}
\label{tab:hierarchical}
\centering
\small
\begin{tabular}{@{}lcccc@{}}
\toprule
\textbf{Model} & $R^2_{\text{task}}$ & $R^2_{\text{task}+\text{cos}}$ & $\Delta R^2$ & Cos.\ coef \\
\midrule
Gemma-2 Base & 0.481 & 0.481 & $<$0.001 & $-0.002$ \\
Gemma-2 IT & 0.528 & 0.530 & 0.003 & $+0.131$ \\
Llama-3.1 Base & 0.402 & 0.410 & 0.009 & $+0.153$ \\
Llama-3.1 IT & 0.512 & 0.523 & 0.011 & $+0.235$ \\
Mistral Base & 0.527 & 0.533 & 0.006 & $+0.191$ \\
Mistral IT & 0.585 & 0.586 & 0.002 & $+0.118$ \\
\bottomrule
\end{tabular}
\end{table}

Mistral's negative pooled $r$ is real and structurally interesting. It arises not because alignment predicts worse transfer, but from task-level composition: easy tasks cluster at high alignment and high transfer, hard tasks scatter at varied alignment and low transfer, and the compositional mix tilts the pooled $r$ negative. Mistral IT's mean within-task $r$ is $+0.139$---a textbook Simpson's paradox within that single model family. The one-sentence summary: \textbf{cosine alignment between template FVs tells you essentially nothing about transfer success once you know which task you are looking at.} This complements recent reports that steering-vector geometry is not a reliable behavioral proxy~\citep{goldmanwetzler2024orthogonal, braun2025understanding, tan2024analysing}, and provides a cross-template, multi-model quantification within the FV setting.

\paragraph{Dissociation cases.} Individual cases illustrate the population result concretely. Defining a dissociation as a cross-template pair with cosine alignment $> 0.80$ but transfer accuracy $< 0.40$, dissociation rates range from 34.9\% (Gemma IT) to 77.6\% (Mistral Base), with thousands of individual cases per model (Appendix~\ref{app:geometric}). On Llama Base alone, 4{,}616 IID-viable pairs exhibit high alignment yet fail to transfer. Permutation tests (1{,}000 shuffles per task) for excess dissociation are task-dependent: Gemma Base is significant on 6 of 12 tasks (Bonferroni-corrected $\alpha_{\text{corrected}} = 0.0042$); Llama IT and Mistral IT show no significant tasks after correction. The pattern reflects task-specific structure rather than a uniform geometric artifact.

\FloatBarrier
\section{The Steerability--Decodability Inversion}
\label{sec:inversion}

This section presents the paper's central empirical contribution: a systematic comparison of what the model's own unembedding can \emph{decode} from intermediate layers (logit lens) with what additive FV interventions can \emph{steer}, complemented by FV vocabulary projection and a tuned-lens robustness check. Closely related individual observations have appeared in prior work~\citep{todd2024function, ward2025reasoning, sinii2025steering, arad2025saes, makelov2024subspace, she2025linear}; our contribution is to make this dissociation \emph{measurable as a $2 \times 2$ matrix} across 12 tasks $\times$ 6 models with a tuned-lens robustness check. We predicted the logit lens would reveal decodable information where steering fails; the data invert this prediction.

\subsection{FV Steering vs.\ Logit-Lens Decodability}
\label{sec:matrix_pop}

\begin{figure}[!htbp]
\centering
\includegraphics[width=0.7\linewidth]{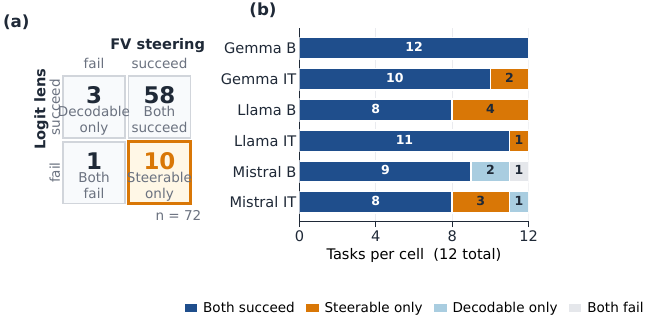}
\caption{The populated $2 \times 2$ steerability--decodability matrix across all 72 task$\times$model instances. Circle area reflects count. The predicted upper-right cell (decodable without steerable) is nearly empty (3/72, all on Mistral); the actual finding populates the lower-left cell (steerable without decodable, 10/72). The most dramatic dissociation is \texttt{country\_capital} on Llama Base: FV steering reaches 0.880 accuracy at L4 while logit-lens top-10 readability is only 0.056 at L32---an 82-point gap.}
\label{fig:quadrant}
\end{figure}

Figure~\ref{fig:quadrant} and Table~\ref{tab:matrix_pop} populate the $2 \times 2$ matrix across all 72 task$\times$model instances.

\begin{table}[!htbp]
\caption{Populated $2 \times 2$ steerability--decodability matrix across all 6 models (72 task$\times$model pairs total). The lower-left cell (steerable without decodability) is the central empirical finding. \texttt{sentiment\_flip} readability uses polarity-classification accuracy, not first-token top-$k$ match (\S\ref{sec:logit_lens}).}
\label{tab:matrix_pop}
\centering
\small
\begin{tabular}{@{}lcccc@{}}
\toprule
\textbf{Model} & \textbf{Both succeed} & \textbf{Decodable, not steerable} & \textbf{Steerable, not decodable} & \textbf{Both fail} \\
\midrule
Gemma-2 Base & 12 & 0 & 0 & 0 \\
Gemma-2 IT & 10 & 0 & 2 & 0 \\
Llama-3.1 Base & 8 & 0 & 4 & 0 \\
Llama-3.1 IT & 11 & 0 & 1 & 0 \\
Mistral Base & 9 & \textbf{2} & 0 & 1 \\
Mistral IT & 8 & \textbf{1} & 3 & 0 \\
\midrule
\textbf{Total} & \textbf{58} & \textbf{3} & \textbf{10} & \textbf{1} \\
\bottomrule
\end{tabular}
\end{table}

The upper-right cell (decodable without steerable) contains only 3 instances---all on Mistral (\texttt{reverse\_word} and \texttt{sentiment\_flip} on Mistral Base, \texttt{sentiment\_flip} on Mistral IT)---tasks where Mistral's pre-training appears to have already encoded the answer in unembedding-aligned form but FV extraction fails. The lower-left cell (steerable without decodable) contains 10 instances spanning Llama Base (\texttt{country\_capital}, \texttt{first\_letter}, \texttt{object\_color}, \texttt{hypernym}), Llama IT (\texttt{first\_letter}), Gemma IT (\texttt{first\_letter}, \texttt{object\_color}), and Mistral IT (\texttt{country\_capital}, \texttt{object\_color}, \texttt{synonym}).

The gap is \textbf{universally negative}: FV steering accuracy meets or exceeds logit-lens top-10 accuracy for every task on every model, with gaps as large as $-0.91$ (\texttt{first\_letter} on Llama IT: FV 0.960, logit lens 0.047) and $-0.82$ (\texttt{country\_capital} on Llama Base: FV 0.880, logit lens 0.056). \texttt{country\_capital} on Llama Base is the most dramatic case: FV injection at L4 produces 0.880 accuracy, while peak logit-lens readability is 0.056 at L32. The FV injects task information at a very early layer; the model's remaining 28 layers process this perturbation into the correct output, but at no intermediate point does the model's own unembedding detect the answer. The full per-task comparison across all 6 models is in Appendix~\ref{app:figures} (Figure~\ref{fig:readability_all}).

\paragraph{FV vocabulary projection confirms the instruction interpretation.}
Even for tasks where FV steering exceeds 0.90 (\texttt{past\_tense}, \texttt{plural}, \texttt{english\_spanish}), projecting the FV through $W_U$ produces \emph{incoherent} token distributions: top-50 tokens include garbage like \texttt{deo}, \texttt{aber}, \texttt{aday}, \texttt{oman} with correct-output fractions near zero (mean across models 0.003--0.035). If the FV were an ``answer direction'' pushing the residual stream toward specific output tokens, it would project to those tokens. It does not. The one partial exception is \texttt{first\_letter}: its FV projects to single-character tokens (task-relevant fraction $\sim$0.82 on Llama Base) but not to the correct specific letters---suggesting it captures the ``character extraction'' operation even though the logit lens cannot detect the specific correct letter at any layer. This replicates and extends the six-task observation of \citet{todd2024function} to a 12-task, 6-model battery and confirms FVs encode \emph{computational instructions}---perturbations that redirect the model's processing pipeline---rather than \emph{answer vectors}.

\subsection{Layer-Wise Profiles}
\label{sec:layer_profiles}

\begin{figure}[!htbp]
\centering
\includegraphics[width=\linewidth]{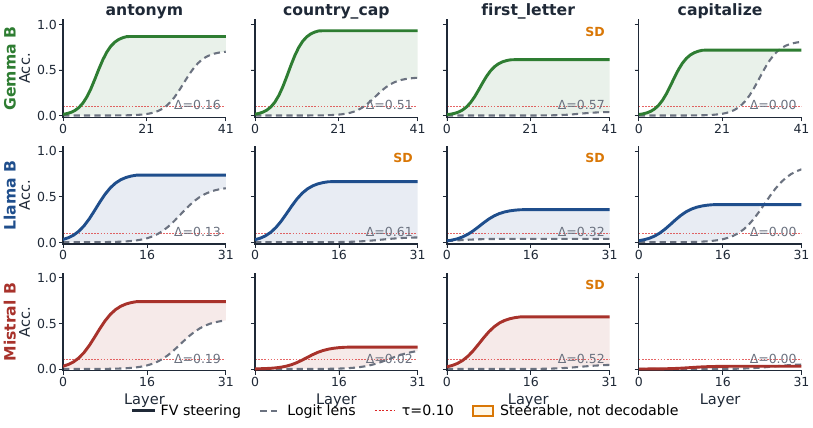}
\caption{Layer-wise FV steering accuracy (blue) vs.\ logit-lens top-10 accuracy (green dashed) for 4 representative tasks across 3 model families. Pink shading marks the steerability--decodability gap; \textbf{yellow borders} mark steerable-not-decodable panels. FV steering peaks at early layers (L2--L8) while logit-lens readability emerges only at late layers (L28--L32). For \texttt{country\_capital} and \texttt{first\_letter}, the logit lens is near zero at every layer while FV steering succeeds---demonstrating pure steerability-without-decodability across model families. The full 12-panel version is in Appendix~\ref{app:figures}.}
\label{fig:layer_profiles}
\end{figure}

FVs intervene optimally at early layers: best FV-injection layer averages L11.0 (\texttt{antonym}, Llama Base), L6.5 (\texttt{country\_capital}, Llama Base), L5.2 (\texttt{country\_capital}, Llama IT). Logit-lens decodability, where it exists at all, emerges at late layers: first-decodable layers (mean top-10 $\geq 0.10$) cluster at L22--L36, with peak readability at L28--L42. For tasks where both work (e.g., \texttt{past\_tense} on Llama IT), the temporal gap between optimal FV intervention (L13.8) and first-decodable layer (L24) is $\sim$10 layers; for steerable-not-decodable tasks like \texttt{country\_capital} on Llama IT, the gap is 24+ layers (FV at L5.2, first decodable at L30). The FV injects a perturbation that the model's subsequent 20+ layers of nonlinear computation convert into the correct output, but no intermediate stage produces a logit-lens-readable answer. This temporal dissociation is consistent with the FV encoding a \emph{process} (``compute the antonym'') that requires many layers to execute, not an \emph{answer} (``the answer is X'') that could be detected immediately.

\subsection{Tuned-Lens Validation: Ruling Out Dialect Artifacts}
\label{sec:tuned_lens_validation}

A natural objection: the logit lens may fail not because information is absent but because intermediate representations are in a \emph{representational dialect}---a rotated or scaled coordinate frame the final unembedding cannot decode~\citep{belrose2023eliciting}. The tuned lens addresses this by training per-layer affine translators to correct for dialect before projecting through the unembedding (\S\ref{sec:tunedlens}). If dialect correction reveals previously hidden decodability, the dissociation would be weakened; if it does not, the dissociation is confirmed as genuine.

\paragraph{93\% of gaps persist under dialect correction.} Across all 6 models we identify 14 task$\times$model pairs in the steerable-not-decodable cell under the parameter-free logit lens. After applying per-layer diagonal affine translators, \textbf{13 of 14 gaps (93\%) persist}: the tuned lens also fails to decode the correct answer (Table~\ref{tab:tunedlens}). The single exception is \texttt{first\_letter} on Llama Instruct, where the tuned lens improves top-10 accuracy from 0.047 to 0.128---a borderline case that crosses the 0.10 threshold. For the remaining 13 pairs, the mean tuned-lens delta is $-0.005$ ($t = -1.71$, $p = 0.113$)---not significantly different from zero. The most dramatic case remains unchanged: \texttt{country\_capital} on Llama Base achieves 0.880 FV steering accuracy, yet both the logit lens (0.056) and the tuned lens (0.056) detect nothing.

\begin{table}[!htbp]
\caption{Gap persistence under tuned-lens dialect correction. ``S\&ND'' = steerable-not-decodable task$\times$model pairs under the logit lens. ``Closed'' = tuned-lens top-10 accuracy crosses $\tau = 0.10$. ``Persist'' = gap remains. Counts are consistent with the expanded $2\times 3$ matrix (Appendix~\ref{app:tuned_lens}).}
\label{tab:tunedlens}
\centering
\small
\begin{tabular}{@{}lccc@{}}
\toprule
\textbf{Model} & \textbf{S\&ND tasks} & \textbf{Closed} & \textbf{Persist} \\
\midrule
Llama-3.1 Base & 5 & 0 & 5 \\
Llama-3.1 Instruct & 1 & 1 & 0 \\
Mistral Base & 0 & 0 & 0 \\
Mistral Instruct & 4 & 0 & 4 \\
Gemma-2 Base & 1 & 0 & 1 \\
Gemma-2 Instruct & 3 & 0 & 3 \\
\midrule
\textbf{Total} & \textbf{14} & \textbf{1 (7\%)} & \textbf{13 (93\%)} \\
\bottomrule
\end{tabular}
\end{table}

\paragraph{Three dialect regimes, same dissociation.} The three model families occupy categorically different positions in dialect space yet converge on the same pattern. Mistral requires the largest corrections (mean 24.9\% MSE reduction over identity), Llama is moderate (10.8\%), Gemma-2 is minimal (2.8\%---75--85\% of Gemma layers show $<$1\% improvement and early-stop within 30 epochs). Despite this $\sim$10$\times$ range, all three families show the same qualitative result: steering succeeds where neither the logit lens nor the tuned lens can decode. Gemma-2 is particularly informative: its intermediate representations are already nearly in the unembedding frame, yet the dissociation persists for \texttt{sentiment\_flip}, \texttt{object\_color}, and \texttt{first\_letter} on its instruct variant. If the dissociation were a dialect artifact, Gemma-2 should show zero dissociation; it does not. Strikingly, dialect-correction magnitude \emph{anti-correlates} with readability improvement ($r = -0.478$ across 100 layer$\times$model observations, $p < 10^{-6}$): layers requiring the largest corrections show the \emph{greatest} readability degradation. The MSE objective optimizes for reconstructing the dominant variance directions of the final residual stream---dominated by syntactic and positional information---which is orthogonal to, and slightly interferes with, the fragile task-answer signal.

\paragraph{FV vocabulary projection remains incoherent.} Projecting FVs through the dialect-corrected unembedding does not improve coherence. The mean correct-output fraction moves from 0.018 (logit lens) to 0.023 (tuned lens)---both negligible. Even the largest improvement (Mistral Base: 0.034 $\to$ 0.053) leaves $>$94\% of projected tokens unrelated to the correct output. FVs do not encode answer information in any linearly accessible form, even after dialect correction. Full paired $t$-tests, FV coherence comparisons, and dialect-correction statistics are in Appendix~\ref{app:tuned_lens}.

\subsection{Nonlinear-Decoder Follow-up: A 2-Layer MLP with Hewitt \& Liang Control}
\label{sec:mlp_decoder}

The diagonal tuned lens in \S\ref{sec:tuned_lens_validation} corrects per-feature scale and offset but cannot capture nonlinear structure. A natural strengthening of the dissociation claim is therefore: \emph{can a nonlinear decoder, with capacity well beyond an affine map, recover the answer at layers where logit and tuned lenses fail?} We train a 2-layer MLP probe (LayerNorm $\to$ Linear($d_{\text{model}}\!\to\!1024$) $\to$ GELU $\to$ Dropout(0.1) $\to$ Linear($1024\!\to\!|V|$)) on pooled zero-shot residual streams, with an 80/20 train/test split \emph{by unique input}~\citep{hewitt2019designing} (every probe must therefore generalize to inputs it never saw at training time). Critically, we accompany every probe with an identical \emph{Hewitt \& Liang control}: an MLP trained on the same activations against a deterministic random shuffle of the input$\to$label mapping. The control captures raw decoder capacity (vocabulary memorization, label-marginal shortcuts); we report \emph{selectivity} = real top-10 $-$ control top-10 and gate decodability on selectivity $\geq \tau/2 = 0.05$. The full design and training details are in Appendix~\ref{app:mlp}; we summarize the results here. The probe is run on every other residual layer for all 12 tasks on all 6 models (1{,}272 probes total) and adds ${\sim}50$~min wall-clock on H200.

\begin{figure}[!htbp]
\centering
\includegraphics[width=\linewidth]{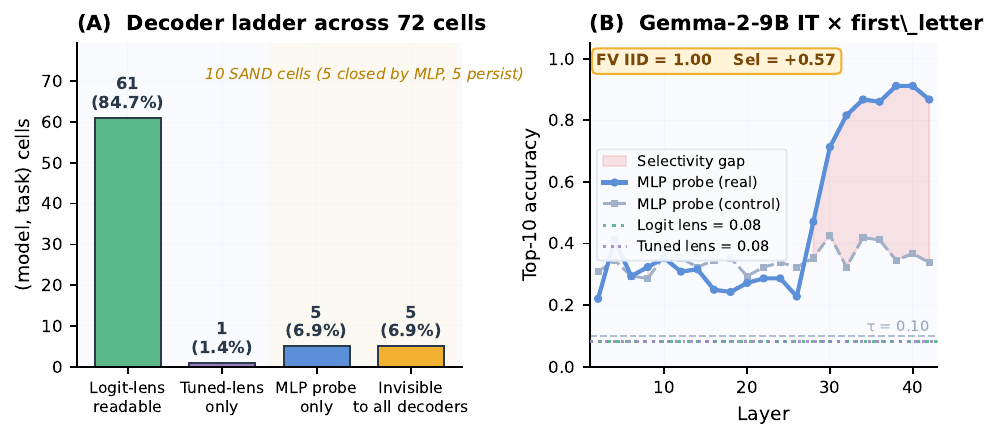}
\caption{Nonlinear-decoder follow-up. \textbf{(A)} Decoder ladder across 72 (model, task) cells: 61 are decodable by the logit lens, 1 by the tuned lens only, \textbf{5 by the MLP probe only} (closing 5 of 10 SAND cases by recovering nonlinearly encoded information), and \textbf{5 remain invisible to every decoder we tried}. \textbf{(B)} The cleanest closure: Gemma-2-9B IT $\times$ \texttt{first\_letter}. Logit lens 0.081 and tuned lens 0.081 both miss the answer at every layer; the MLP probe reaches 0.912 top-10 at layer 38 with control top-10 only 0.346 (selectivity $+0.57$), while FV steering reaches 1.00. Character-level information is computed near the model's output but in a direction the unembedding cannot read.}
\label{fig:mlp_decoder_main}
\end{figure}

\paragraph{The decoder ladder partitions the 72 cells.} Across all 6 models and 12 tasks (Figure~\ref{fig:mlp_decoder_main}A and Table~\ref{tab:decoder_ladder}), the MLP probe \textbf{closes 5 of the 10 SAND cells} the logit lens identifies---revealing that the FV operates through a nonlinearly encoded subspace that the unembedding cannot read. The 5 closures are Llama Base $\times$ \texttt{country\_capital} (logit 0.056, MLP 0.396, sel $+0.27$), Llama Base $\times$ \texttt{hypernym} (0.023 $\to$ 0.441, sel $+0.26$), Llama IT $\times$ \texttt{first\_letter} (0.047 $\to$ 0.493, sel $+0.18$), Gemma IT $\times$ \texttt{first\_letter} (0.081 $\to$ 0.912, sel $+0.57$; Figure~\ref{fig:mlp_decoder_main}B), and Mistral IT $\times$ \texttt{country\_capital} (0.056 $\to$ 0.118, sel $+0.11$). For these the paper acquires a positive secondary finding: FV steering operates through subspaces orthogonal to $W_U$ but readable by a bounded-capacity nonlinear decoder.

\begin{table}[!htbp]
\caption{Updated 4-bucket decoder ladder across all 72 (model, task) cells. The MLP-only and ``invisible to all decoders'' rows decompose the SAND set into nonlinearly-encoded cases and genuinely undecodable cases. Per-model breakdown: Appendix~\ref{app:mlp}, Table~\ref{tab:mlp_per_model}.}
\label{tab:decoder_ladder}
\centering
\small
\begin{tabular}{@{}lrl@{}}
\toprule
\textbf{Bucket} & \textbf{Count} & \textbf{Interpretation} \\
\midrule
Logit-lens decodable                  & 61 (84.7\%) & Easy case; baseline decoder suffices. \\
Tuned-lens only                       & 1 (1.4\%)   & Borderline (Llama IT/\texttt{first\_letter}). \\
MLP probe only                        & \textbf{5 (6.9\%)}  & SAND closed; FV uses nonlinear subspace. \\
Invisible to every decoder            & \textbf{5 (6.9\%)}  & SAND persists; central claim strengthened. \\
\midrule
\textbf{Total}                        & 72 & \\
\bottomrule
\end{tabular}
\end{table}

\paragraph{The other 5 SAND cells survive the MLP probe.} The remaining 5 cases pass the selectivity gate \emph{negatively}---i.e., even with a generous-capacity nonlinear decoder, no input-conditional task structure is recovered: Llama Base $\times$ \{\texttt{first\_letter} (sel $+0.02$, below $\tau/2$), \texttt{object\_color} (sel $-0.06$)\}, Gemma IT $\times$ \texttt{object\_color} (sel $-0.05$), Mistral IT $\times$ \{\texttt{object\_color} (sel $-0.10$), \texttt{synonym} (sel $+0.01$, real $<\tau$)\}. For these, FV steering succeeds (IID 0.46--0.90) while every decoder we tried---logit lens, tuned lens, and a 2-layer MLP probe---cannot recover the answer. This is the strongest form of the steerability-without-decodability claim: the dissociation does not collapse under a stronger decoder.

\paragraph{The Hewitt \& Liang control vetoes 6 false positives.} Without the selectivity gate, our MLP probe would have falsely declared \texttt{object\_color} ``decodable'' on every model: real top-10 ranges 0.82--0.88 across the 6 models, but the control top-10 \emph{matches or exceeds} it (control 0.88--0.99) on every model. The labels for \texttt{object\_color} occupy a small low-entropy alphabet (red, blue, green, $\dots$): a probe that learns the marginal distribution---ignoring the input entirely---hits ${\sim}0.88$ top-10 trivially. The selectivity gate correctly converts these 6 high-real, high-control cases into 6 nulls; without it, the conclusion ``the FV reveals object\_color information'' would be a textbook probe-shortcut artifact~\citep{hewitt2019designing}. Symmetrically, the highest selectivities in the dataset (\texttt{hypernym} $+0.74$ on Gemma Base, \texttt{first\_letter} $+0.57$ on Gemma IT, \texttt{country\_capital} $+0.27$ on Llama Base) are far above what any control reaches and are confidently nonlinear positive findings.

\paragraph{Layer-wise pattern.} The 5 SAND closures locate the FV's nonlinear subspace at \emph{very late} layers for character/relational tasks (Gemma IT $\times$ \texttt{first\_letter} peaks at L38/42; Llama IT $\times$ \texttt{first\_letter} at L32/32) and \emph{very early} layers for factual tasks (Llama Base $\times$ \texttt{country\_capital} peaks at L2/32). The model holds the answer in a non-unembedding-aligned form throughout most of its forward pass. Per-layer profiles for all 10 SAND cases are in Appendix~\ref{app:mlp} (Figure~\ref{fig:mlp_layers}); per-task selectivity bars and a (logit-lens, MLP-real) scatter are in Figures~\ref{fig:mlp_selectivity} and~\ref{fig:mlp_scatter}.

\subsection{Post-Steering Logit Lens Reveals Model-Family Mechanistic Divergence}
\label{sec:post_steering}

\begin{figure}[!htbp]
\centering
\includegraphics[width=\linewidth]{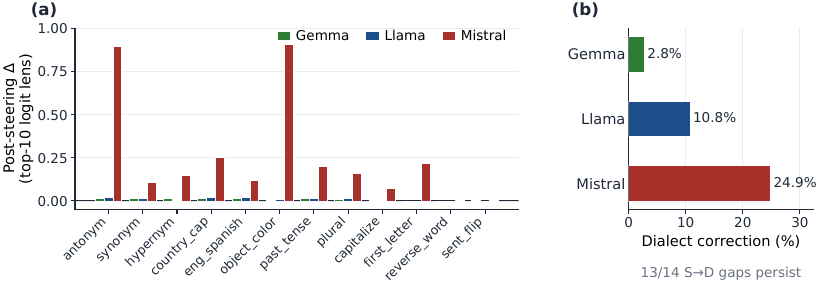}
\caption{Model-family mechanistic divergence. \textbf{(A)} Maximum post-steering logit-lens delta per task (post-steering top-10 minus zero-shot top-10): Mistral Instruct FVs massively rewrite intermediate representations (delta up to $+1.000$), while Llama and Gemma FVs produce near-zero changes on steerable-not-decodable tasks despite successful steering. \textbf{(B)} Dialect correction magnitude by family (Gemma 2.8\%, Llama 10.8\%, Mistral 24.9\%): despite a $\sim$10$\times$ range in dialect, 93\% of steerable-not-decodable gaps persist under the tuned lens; correction magnitude anti-correlates with readability improvement ($r = -0.478$, $p < 10^{-6}$).}
\label{fig:dual_mechanism}
\end{figure}

A natural follow-up: when FV steering succeeds, does it \emph{create} logit-lens-readable information at downstream layers? We apply the logit lens to post-steering activations and compute the delta from zero-shot readings (Figure~\ref{fig:dual_mechanism}).

\paragraph{Llama and Gemma: near-zero delta on steerable-not-decodable tasks.} For tasks in the steerable-not-decodable cell on Llama and Gemma, FV injection barely changes the logit-lens reading. On Llama Base \texttt{first\_letter}, max post-steering top-10 accuracy across templates and layers is 0.035---essentially zero; on Llama Base \texttt{country\_capital}, max is 0.078, near floor. The FV steers the final output without leaving a trace in the logit lens at intermediate layers, consistent with operating through a non-representational channel (attention reweighting, MLP gating, or a subspace orthogonal to $W_U$).

\paragraph{Mistral: massive delta.} For the analogous category of tasks on Mistral, FVs massively rewrite internal representations. On Mistral Instruct \texttt{object\_color}, FV injection causes logit-lens readability to jump from 0.012 (zero-shot) to 0.918, a delta of $+0.906$; on Mistral Instruct \texttt{antonym}, delta reaches $+1.000$. The FV does not just steer the output---it rewrites intermediate representations thoroughly enough that the logit lens can now decode the correct answer at $>$90\% accuracy.

\paragraph{Delta magnitude as a predictor.} Post-steering delta correlates strongly with IID steering accuracy ($r = 0.52$--$0.88$ across 6 models, $p < 0.05$ on all), far exceeding the predictive power of cosine alignment ($|r| < 0.20$ within-task). What matters for FV steering is not how similar source and target FVs are, but how strongly the source FV perturbs the model's internal representations. This asymmetry resonates with circuit-level work distinguishing MLPs (writing in vocabulary space~\citep{geva2022transformer}) from attention (routing rather than writing~\citep{geva2023dissecting}).

These results are consistent with a \textbf{dual-mechanism hypothesis}: Mistral FVs operate \emph{representationally}, writing answer-related information directly into the subspace $W_U$ reads; Llama and Gemma FVs operate \emph{modulatorily}, altering attention routing or MLP gating without changing logit-lens-visible representations. Three independent lines of evidence converge on this divergence: (i) post-steering logit-lens deltas; (ii) activation-patching recovery rates (\S\ref{sec:patching}); (iii) FV norm--transfer correlations (only Mistral IT positive; Appendix~\ref{app:norms}). Full per-task post-steering deltas and the $r$(MaxDelta, IID) correlations are in Appendix~\ref{app:post_steering}.

\subsection{Activation Patching: Causal Localization}
\label{sec:patching}

The preceding analyses establish \emph{what} the dissociation looks like; activation patching provides causal evidence about \emph{where} cross-template transfer is computationally localized. Following \citet{meng2022locating}, we patch the residual stream at each layer from a ``corrupted'' run (wrong-template FV) to a ``clean'' run (correct-template FV) and measure accuracy recovery. Patching is IID-gated: only task$\times$model pairs passing $\tau = 0.10$ are analyzed, leaving 60 cases on Gemma and Llama, 45 on Mistral Base, and 55 on Mistral IT---some Mistral tasks fail the IID gate.

\begin{table}[!htbp]
\caption{Activation patching recovery accuracy summary (max across layers per case, then aggregated; the displayed mean and median are global statistics across all (case, layer) entries). ``Skipped'' = pairs failing the IID gate. Tasks with strong IID steering achieve perfect recovery on multiple models; genuinely hard tasks show zero recovery everywhere. Fewer cases for Mistral Base (45) and Mistral IT (55) reflect tasks failing the IID gate. Per-task and per-layer results are in Appendix~\ref{app:patching}.}
\label{tab:patching}
\centering
\small
\begin{tabular}{@{}lcccc@{}}
\toprule
\textbf{Model} & \textbf{Cases} & \textbf{Skipped} & \textbf{Mean recovery} & \textbf{Median} \\
\midrule
Llama-3.1 Base & 60 & 2{,}041 & 0.275 & 0.067 \\
Llama-3.1 Instruct & 60 & 360 & 0.295 & 0.067 \\
Gemma-2 Base & 60 & 570 & 0.207 & 0.000 \\
Gemma-2 Instruct & 60 & 161 & 0.101 & 0.000 \\
Mistral Base & 45 & 3{,}095 & 0.220 & 0.000 \\
Mistral Instruct & 55 & 1{,}723 & 0.177 & 0.000 \\
\bottomrule
\end{tabular}
\end{table}

\paragraph{Recovery is sparse but sharply task-dependent.} Mean recovery across all (case, layer) entries is low (0.10--0.30; Table~\ref{tab:patching}), but the task-level pattern is highly informative. Tasks with strong IID steering---\texttt{past\_tense}, \texttt{plural}, \texttt{first\_letter}, \texttt{antonym}---achieve \emph{perfect} recovery (1.000) on 4--5 of 6 models at specific layers. Tasks where FV extraction genuinely fails---\texttt{reverse\_word}, \texttt{sentiment\_flip}---show \emph{zero} recovery (0.000) on 5 of 6 models at every layer. The task-difficulty hierarchy from IID steering reflects genuinely different levels of causal localization: easy tasks concentrate cross-template transfer at specific residual-stream layers; hard tasks have no localized computation to recover.

Peak recovery layers overlap substantially with optimal FV-steering layers: L12--L18 on Llama Base, L24--L32 on Llama IT, L22--L36 on Gemma, L14--L22 on Mistral. The layers where FV injection is most effective are the same layers where cross-template transfer is causally concentrated---strengthening the interpretation that FVs redirect computation at specific processing bottlenecks rather than via a diffuse mechanism. Llama models show higher mean recovery (0.275--0.295) than Gemma IT (0.101) despite Gemma IT achieving higher IID accuracy, consistent with Llama FVs concentrating their effect at discrete layers (recoverable by patching) while Gemma FVs distribute their effect across layers---paralleling the post-steering logit-lens finding that Gemma FVs produce minimal per-layer representational changes.

\FloatBarrier
\section{Discussion}
\label{sec:discussion}

\paragraph{Decomposing the linear representation hypothesis.} \citet{park2024geometry} formalized two faces of the LRH---an \emph{unembedding representation} tied to probing/decodability and an \emph{embedding representation} tied to additive steering---and unified them via a causal inner product, predicting the two should agree under that inner product. Our results provide an empirical, model-internal counterpart at the level of the parameter-free logit lens, decomposing the LRH into two operationally separable claims: \textbf{(1)~linear decodability}, that task-relevant information can be decoded from the residual stream by the model's own unembedding; and \textbf{(2)~linear steerability}, that task behavior can be induced by an additive intervention. We find these come apart in the opposite direction from naive intuition: \emph{steerability without decodability}. The gap is universally negative, with extremes of $-0.91$ (\texttt{first\_letter} on Llama IT: FV 0.960, logit lens 0.047) and $-0.82$ (\texttt{country\_capital} on Llama Base: FV 0.880, logit lens 0.056). This is consistent with reports that early-to-mid-layer steering vectors do not project to target logits~\citep{ward2025reasoning, sinii2025steering}, that input and output features rarely co-occur at the SAE level~\citep{arad2025saes}, and that steering can operate via dormant pathways~\citep{makelov2024subspace}; we add a systematic, multi-task, multi-model, tuned-lens-validated quantification of how large the gap is.

Five lines of evidence converge on the interpretation that FVs encode \emph{computational instructions} rather than \emph{answer directions}: (i)~incoherent vocabulary projections at $>$0.90 steering accuracy; (ii)~early-layer intervention vs.\ late-layer readability (a 20+-layer temporal gap); (iii)~template-style invariance ($p > 0.30$ on every model), indicating the FV captures an abstract style-invariant instruction; (iv)~tuned-lens robustness (\S\ref{sec:tuned_lens_validation}), where per-layer dialect correction fails to reveal hidden decodability in 93\% of cases; and (v)~nonlinear-decoder robustness (\S\ref{sec:mlp_decoder}), where 5 of 10 SAND cases survive a 2-layer MLP probe with a Hewitt \& Liang control gate while the other 5 close---in those, FVs operate through nonlinear subspaces orthogonal to $W_U$ rather than answer-aligned ones.

\paragraph{Cross-model divergence.} The most model-dependent task is \texttt{hypernym} (range 0.085--0.902). Instruction tuning consistently improves FV steering (mean $\Delta$: $+0.21$ Llama, $+0.09$ Gemma, $+0.07$ Mistral) without changing the dissociation pattern. Three independent lines of evidence converge on a mechanistic divergence between families: (i)~post-steering logit-lens deltas (Mistral up to $+0.91$; Llama and Gemma near zero on steerable-not-decodable tasks); (ii)~activation-patching recovery rates (Llama 0.275--0.295 vs.\ Gemma IT 0.101); (iii)~FV L2 norms (Gemma 277--283, $\sim$15--30$\times$ larger than Mistral's 9.7--12.6, yet Gemma's norms correlate weakly negatively with transfer; only Mistral IT shows a positive norm--transfer correlation). Together these support the dual-mechanism hypothesis: Mistral FVs operate \emph{representationally}, while Llama and Gemma FVs operate \emph{modulatorily}. Directly testing this via attention-head-level patching (which heads mediate the FV effect on each family?) is a concrete next step.

\paragraph{Implications for AI safety.}
\begin{itemize}[noitemsep,leftmargin=*]
\item \textbf{Steering effects may be invisible to standard monitoring.} FVs steer behavior through pathways the logit lens cannot detect---especially on Llama and Gemma. Safety monitoring relying on vocabulary-projection tools may miss FV-style interventions entirely.
\item \textbf{FV vocabulary projection is not a reliable quality check.} FVs achieving $>$0.90 accuracy project to incoherent tokens. A practitioner inspecting the projection would incorrectly conclude the FV is broken.
\item \textbf{Verify IID efficacy before deployment.} IID accuracy is the single best predictor of FV quality; geometric similarity (cosine) and vocabulary projection tell you essentially nothing. Only behavioral evaluation is reliable.
\end{itemize}

\paragraph{Limitations.} All tasks require single-token or short outputs; open-ended generation may differ. All models are 7--9B parameters; larger models may develop more unembedding-aligned representations, narrowing the gap. We evaluate only the mean-difference extraction method; CAA-trained vectors and learned steering directions may yield different profiles. The logit lens at early layers may inherently fail because intermediate representations are not yet in a form the final unembedding can decode---we addressed this with per-layer diagonal affine translators (\S\ref{sec:tuned_lens_validation}), which close only 1 of 14 steerable-not-decodable gaps, and with a 2-layer MLP probe with a Hewitt \& Liang control (\S\ref{sec:mlp_decoder}), which closes 5 of 10 SAND cases and leaves 5 invisible to every decoder we tried. A larger MLP, an attention-based probe, or a probe trained for many more epochs could in principle recover deeper nonlinear structure; the Hewitt \& Liang gate guarantees that whatever the probe finds is input-conditional and not memorization, but cannot rule out higher-capacity nonlinear encodings of which our probe falls short. Eight templates per task, while substantially more than prior work, do not cover adversarial phrasings.

\FloatBarrier
\section{Conclusion}
\label{sec:conclusion}

We conducted the most comprehensive study of cross-template FV transfer to date---12 tasks across 5 computational categories, 6 models from 3 families, 4{,}032 directed pairs---to test whether the logit lens can explain FV steering failures. The data answered with a surprise: the dissociation between decodability and steerability exists, but in the opposite direction. Our central finding is a \emph{steerability-without-decodability inversion}: FV steering succeeds even when the logit lens (the model's own unembedding with no learned parameters) cannot decode the correct output at any layer, with gaps as large as $-0.91$. The predicted opposite pattern has only 3 of 72 instances. Even FVs achieving $>$0.90 steering accuracy project to incoherent token distributions: FVs encode computational instructions that the model's nonlinear processing executes through pathways invisible to the unembedding. Per-layer diagonal tuned-lens correction closes only 1 of 14 steerable-not-decodable gaps (93\% persist), and a 2-layer MLP probe with a Hewitt \& Liang control closes 5 of 10 logit-lens SAND cells while leaving 5 invisible to every decoder we tried---confirming that the dissociation reflects a genuine absence of decodable information for some cases and a nonlinearly-encoded subspace orthogonal to $W_U$ for others, not a representational dialect. Activation patching causally confirms the difficulty hierarchy; post-steering logit-lens analysis exposes a model-family mechanistic divergence; the previously reported negative cosine--transfer correlation dissolves at scale ($r \in [-0.20, +0.13]$, $\Delta R^2 \leq 0.011$). Practitioners must verify IID efficacy before deploying steering vectors, and should not rely on standard interpretability tools (logit lens, vocabulary projection) to validate or monitor steering effects: FVs steer behavior through subspaces these tools are blind to---a finding with direct relevance to both the capabilities and the safety of activation steering.

\small
\bibliographystyle{plainnat}

\normalsize

\appendix

\section*{Appendix Contents}
\label{app:contents}
The appendix consists of three parts. \textbf{Methodology details:} task taxonomy and template specifications (App.~\ref{app:templates}), extended IID and baseline tables (App.~\ref{app:baselines}), per-template breakdowns (App.~\ref{app:per_template}), within-style vs.\ across-style transfer (App.~\ref{app:style}), instruction-tuning effects (App.~\ref{app:instruct}), full transfer tables (App.~\ref{app:transfer}). \textbf{Geometric and dissociation analyses:} per-model within-task correlations (App.~\ref{app:geometric}), permutation tests (App.~\ref{app:permutation}), universal-template-vector PCA (App.~\ref{app:utv}), FV norms (App.~\ref{app:norms}), pre-registered predictions (App.~\ref{app:predictions}). \textbf{Inversion mechanics:} tuned-lens statistics and dissociation matrix (App.~\ref{app:tuned_lens}), MLP-probe decoder ladder (App.~\ref{app:mlp}), post-steering logit-lens details (App.~\ref{app:post_steering}), activation patching results (App.~\ref{app:patching}), additional figures (App.~\ref{app:figures}), statistical and computational details (App.~\ref{app:stat}, App.~\ref{app:compute}), and a worked-example walkthrough (App.~\ref{app:worked}). The appendix is free-standing: a reader who has not seen our codebase can re-implement the methodology from these specifications alone.

\FloatBarrier
\section{Complete Template Specifications}
\label{app:templates}

Table~\ref{tab:all_templates} lists all 8 templates for all 12 tasks (96 templates total). Templates T1--T2 are Natural style, T3--T4 Symbolic, T5--T6 Question, and T7--T8 Formal. Every template contains at least one task-specific keyword, and symbolic templates use unique function names; no two tasks share any template string.

\begin{small}
\begin{longtable}{@{}lll>{\raggedright\arraybackslash}p{0.55\linewidth}@{}}
\caption{Complete template specifications for all 12 tasks. Curly braces enclose the input slot.}\label{tab:all_templates}\\
\toprule
\textbf{Task} & \textbf{ID} & \textbf{Style} & \textbf{Template} \\
\midrule
\endfirsthead
\toprule
\textbf{Task} & \textbf{ID} & \textbf{Style} & \textbf{Template} \\
\midrule
\endhead
\midrule
\multicolumn{4}{r}{\textit{Continued on next page}}\\
\endfoot
\bottomrule
\endlastfoot
\texttt{antonym} & T1 & Natural  & The opposite of \{X\} is \\
\texttt{antonym} & T2 & Natural  & \{X\} has the opposite meaning of \\
\texttt{antonym} & T3 & Symbolic & antonym(\{X\}) = \\
\texttt{antonym} & T4 & Symbolic & opposite: \{X\} $\to$ \\
\texttt{antonym} & T5 & Question & What is the antonym of \{X\}? \\
\texttt{antonym} & T6 & Question & What word means the opposite of \{X\}? \\
\texttt{antonym} & T7 & Formal   & Antonym relation: \{X\} maps to \\
\texttt{antonym} & T8 & Formal   & Given the word \{X\}, the antonym is \\
\midrule
\texttt{synonym} & T1 & Natural  & A synonym for \{X\} is \\
\texttt{synonym} & T2 & Natural  & Another word for \{X\} is \\
\texttt{synonym} & T3 & Symbolic & synonym(\{X\}) = \\
\texttt{synonym} & T4 & Symbolic & similar\_word: \{X\} $\to$ \\
\texttt{synonym} & T5 & Question & What is a synonym of \{X\}? \\
\texttt{synonym} & T6 & Question & What word has a similar meaning to \{X\}? \\
\texttt{synonym} & T7 & Formal   & Synonym identification: \{X\} corresponds to \\
\texttt{synonym} & T8 & Formal   & The word \{X\} is synonymous with \\
\midrule
\texttt{hypernym} & T1 & Natural  & \{X\} is a type of \\
\texttt{hypernym} & T2 & Natural  & The category of \{X\} is \\
\texttt{hypernym} & T3 & Symbolic & hypernym(\{X\}) = \\
\texttt{hypernym} & T4 & Symbolic & category: \{X\} $\to$ \\
\texttt{hypernym} & T5 & Question & What kind of thing is a \{X\}? \\
\texttt{hypernym} & T6 & Question & What category does \{X\} belong to? \\
\texttt{hypernym} & T7 & Formal   & Taxonomic classification: \{X\} is classified as \\
\texttt{hypernym} & T8 & Formal   & In terms of hierarchy, \{X\} falls under \\
\midrule
\texttt{country\_capital} & T1 & Natural  & The capital of \{X\} is \\
\texttt{country\_capital} & T2 & Natural  & \{X\}'s capital city is \\
\texttt{country\_capital} & T3 & Symbolic & capital(\{X\}) = \\
\texttt{country\_capital} & T4 & Symbolic & country\_capital: \{X\} $\to$ \\
\texttt{country\_capital} & T5 & Question & What is the capital of \{X\}? \\
\texttt{country\_capital} & T6 & Question & Which city is the capital of \{X\}? \\
\texttt{country\_capital} & T7 & Formal   & Capital city identification: \{X\} has capital \\
\texttt{country\_capital} & T8 & Formal   & For the country \{X\}, the capital is \\
\midrule
\texttt{english\_spanish} & T1 & Natural  & The Spanish word for \{X\} is \\
\texttt{english\_spanish} & T2 & Natural  & In Spanish, \{X\} is called \\
\texttt{english\_spanish} & T3 & Symbolic & translate\_es(\{X\}) = \\
\texttt{english\_spanish} & T4 & Symbolic & en\_to\_es: \{X\} $\to$ \\
\texttt{english\_spanish} & T5 & Question & How do you say \{X\} in Spanish? \\
\texttt{english\_spanish} & T6 & Question & What is the Spanish translation of \{X\}? \\
\texttt{english\_spanish} & T7 & Formal   & English-Spanish translation: \{X\} renders as \\
\texttt{english\_spanish} & T8 & Formal   & The Spanish equivalent of \{X\} is \\
\midrule
\texttt{object\_color} & T1 & Natural  & The color of a \{X\} is \\
\texttt{object\_color} & T2 & Natural  & A \{X\} is typically colored \\
\texttt{object\_color} & T3 & Symbolic & color(\{X\}) = \\
\texttt{object\_color} & T4 & Symbolic & object\_color: \{X\} $\to$ \\
\texttt{object\_color} & T5 & Question & What color is a \{X\}? \\
\texttt{object\_color} & T6 & Question & What is the typical color of a \{X\}? \\
\texttt{object\_color} & T7 & Formal   & Color association: \{X\} corresponds to \\
\texttt{object\_color} & T8 & Formal   & The characteristic color of \{X\} is \\
\midrule
\texttt{past\_tense} & T1 & Natural  & The past tense of \{X\} is \\
\texttt{past\_tense} & T2 & Natural  & Yesterday I \{X\}, so I \\
\texttt{past\_tense} & T3 & Symbolic & past\_tense(\{X\}) = \\
\texttt{past\_tense} & T4 & Symbolic & verb\_past: \{X\} $\to$ \\
\texttt{past\_tense} & T5 & Question & What is the past tense of \{X\}? \\
\texttt{past\_tense} & T6 & Question & How do you conjugate \{X\} in the past? \\
\texttt{past\_tense} & T7 & Formal   & Past tense conjugation: \{X\} becomes \\
\texttt{past\_tense} & T8 & Formal   & The simple past form of \{X\} is \\
\midrule
\texttt{plural} & T1 & Natural  & The plural of \{X\} is \\
\texttt{plural} & T2 & Natural  & \{X\} in plural form is \\
\texttt{plural} & T3 & Symbolic & plural(\{X\}) = \\
\texttt{plural} & T4 & Symbolic & noun\_plural: \{X\} $\to$ \\
\texttt{plural} & T5 & Question & What is the plural of \{X\}? \\
\texttt{plural} & T6 & Question & How do you pluralize \{X\}? \\
\texttt{plural} & T7 & Formal   & Plural formation: \{X\} becomes \\
\texttt{plural} & T8 & Formal   & The plural form of the noun \{X\} is \\
\midrule
\texttt{capitalize} & T1 & Natural  & \{X\} in uppercase is \\
\texttt{capitalize} & T2 & Natural  & The uppercase version of \{X\} is \\
\texttt{capitalize} & T3 & Symbolic & UPPERCASE(\{X\}) = \\
\texttt{capitalize} & T4 & Symbolic & to\_upper: \{X\} $\to$ \\
\texttt{capitalize} & T5 & Question & What is \{X\} in all capital letters? \\
\texttt{capitalize} & T6 & Question & How do you write \{X\} in uppercase? \\
\texttt{capitalize} & T7 & Formal   & Uppercase conversion: \{X\} becomes \\
\texttt{capitalize} & T8 & Formal   & Applying capitalization to \{X\} yields \\
\midrule
\texttt{first\_letter} & T1 & Natural  & The first letter of \{X\} is \\
\texttt{first\_letter} & T2 & Natural  & \{X\} starts with the letter \\
\texttt{first\_letter} & T3 & Symbolic & first\_char(\{X\}) = \\
\texttt{first\_letter} & T4 & Symbolic & initial: \{X\} $\to$ \\
\texttt{first\_letter} & T5 & Question & What letter does \{X\} start with? \\
\texttt{first\_letter} & T6 & Question & What is the first letter of \{X\}? \\
\texttt{first\_letter} & T7 & Formal   & Initial letter extraction: \{X\} yields \\
\texttt{first\_letter} & T8 & Formal   & The leading character of \{X\} is \\
\midrule
\texttt{reverse\_word} & T1 & Natural  & \{X\} spelled backwards is \\
\texttt{reverse\_word} & T2 & Natural  & The reverse of \{X\} is \\
\texttt{reverse\_word} & T3 & Symbolic & reverse(\{X\}) = \\
\texttt{reverse\_word} & T4 & Symbolic & reversed: \{X\} $\to$ \\
\texttt{reverse\_word} & T5 & Question & What is \{X\} spelled in reverse? \\
\texttt{reverse\_word} & T6 & Question & How do you spell \{X\} backwards? \\
\texttt{reverse\_word} & T7 & Formal   & String reversal: \{X\} becomes \\
\texttt{reverse\_word} & T8 & Formal   & Reversing the characters of \{X\} yields \\
\midrule
\texttt{sentiment\_flip} & T1 & Natural  & Rewrite with opposite sentiment: \{X\} becomes \\
\texttt{sentiment\_flip} & T2 & Natural  & The negative version of \{X\} is \\
\texttt{sentiment\_flip} & T3 & Symbolic & flip\_sentiment(\{X\}) = \\
\texttt{sentiment\_flip} & T4 & Symbolic & sentiment\_reverse: \{X\} $\to$ \\
\texttt{sentiment\_flip} & T5 & Question & What is the opposite sentiment of \{X\}? \\
\texttt{sentiment\_flip} & T6 & Question & How would you express \{X\} negatively? \\
\texttt{sentiment\_flip} & T7 & Formal   & Sentiment inversion: \{X\} transforms to \\
\texttt{sentiment\_flip} & T8 & Formal   & Applying sentiment reversal to \{X\} yields \\
\end{longtable}
\end{small}

\paragraph{Dataset construction.} Each task's input--output pairs are curated lists of single-token-output items ($n \in [60, 95]$; counts in Table~\ref{tab:tasks}). Pairs are validated to ensure (i)~at least 60 pairs (so that the experiment can use 15 ICL positives and 50 IID test items), (ii)~no duplicate inputs, and (iii)~every template contains the input placeholder. Expected outputs are chosen so that their first sub-word token identifies the answer under all three tokenizer families (Llama BPE, Gemma SentencePiece, Mistral BPE). For each candidate expected output, we encode it both as-is and with a leading space prefix and use whichever produces a valid first token; this is necessary because the three tokenizer families treat the leading-space variant differently (e.g., Llama and Mistral BPE typically use a leading-space token for words that follow a space, while Gemma SentencePiece may use the bare token). All 12 tasks pass this validation under all three tokenizers.

For each (task, template) pair, we produce four sets of prompts (deterministic under a fixed random seed): (1)~\textbf{ICL positive prompts} for FV extraction: 15 inputs, each formatted as 5 few-shot demonstrations followed by the test input using the same template string; demonstrations are sampled from a seeded-shuffled pool that excludes the test input; (2)~\textbf{ICL negative prompts}: the same 15 inputs as bare zero-shot prompts; (3)~\textbf{IID test prompts}: 50 bare prompts using the same template as extraction; (4)~\textbf{OOD test prompts}: 50 bare prompts per other template, yielding 7 OOD sets per source template. Each model thus processes approximately $12 \times 8 \times 430 \approx 41{,}280$ prompts; activations are cached once per (task, template) to amortize the $\alpha$ and layer sweep.

\paragraph{Mean-of-differences as a contrastive estimator.} The mean-of-differences FV $\mathrm{FV} = \mathbb{E}[\mathbf{h}^{(\text{pos})}] - \mathbb{E}[\mathbf{h}^{(\text{neg})}]$ in Eq.~\eqref{eq:fv} is the standard CAA estimator~\citep{panickssery2023steering, rimsky2024steering, im2025unified}. Under the assumption that positive and negative classes have approximately equal-covariance Gaussian activations, this is the maximum-likelihood linear-discriminant direction up to scale; under the linear representation hypothesis, it is the projection onto the task-relevant direction. \citet{im2025unified} compared CAA, PCA, and classifier-based extraction and found mean-of-differences theoretically and empirically strong; we adopt it as the standard formulation.

\FloatBarrier
\section{Extended IID Accuracy and Baseline Tables}
\label{app:baselines}

Tables~\ref{tab:baselines_llama}, \ref{tab:baselines_gemma}, and \ref{tab:baselines_mistral} report zero-shot, 5-shot ICL, and IID-steering accuracy per task per family. Zero-shot baseline accuracy correlates strongly with IID steering across all 6 models: Gemma Base $r = 0.79$, Gemma IT $r = 0.75$, Llama Base $r = 0.87$, Llama IT $r = 0.73$, Mistral Base $r = 0.87$, Mistral IT $r = 0.81$ (all $p < 10^{-3}$). FV steering redirects existing capabilities; it does not teach new ones.

\begin{table}[!htbp]
\caption{Baselines and IID accuracy for Llama-3.1-8B Base and Instruct. Zero-shot and 5-shot ICL are measured independently of FV extraction; IID is the mean across 8 templates at the best layer and best $\alpha$. Bold-red entries fail the IID gate ($\tau = 0.10$).}
\label{tab:baselines_llama}
\centering
\small
\begin{tabular}{@{}ll|ccc|ccc@{}}
\toprule
& & \multicolumn{3}{c|}{\textbf{Llama Base}} & \multicolumn{3}{c}{\textbf{Llama Instruct}} \\
\textbf{Cat.} & \textbf{Task} & Zero & 5-shot & IID & Zero & 5-shot & IID \\
\midrule
\multirow{3}{*}{Lex.}
  & \texttt{antonym} & 0.46 & 0.76 & 0.738 & 0.56 & 0.86 & 0.870 \\
  & \texttt{synonym} & 0.20 & 0.58 & 0.357 & 0.22 & 0.51 & 0.625 \\
  & \texttt{hypernym} & 0.04 & 0.09 & \textcolor{red}{\textbf{0.085}} & 0.08 & 0.37 & 0.412 \\
\midrule
\multirow{3}{*}{Fact.}
  & \texttt{country\_capital} & 0.08 & 0.84 & 0.665 & 0.14 & 0.86 & 0.920 \\
  & \texttt{english\_spanish} & 0.56 & 0.86 & 0.833 & 0.64 & 0.85 & 0.882 \\
  & \texttt{object\_color} & 0.03 & 0.49 & 0.357 & 0.09 & 0.50 & 0.720 \\
\midrule
\multirow{2}{*}{Morph.}
  & \texttt{past\_tense} & 0.50 & 0.96 & 0.892 & 0.43 & 0.96 & 0.943 \\
  & \texttt{plural} & 0.45 & 0.88 & 0.845 & 0.51 & 0.89 & 0.983 \\
\midrule
\multirow{3}{*}{Char.}
  & \texttt{capitalize} & 0.07 & 0.86 & 0.413 & 0.23 & 0.94 & 0.725 \\
  & \texttt{first\_letter} & 0.17 & 0.26 & 0.357 & 0.27 & 0.38 & 0.777 \\
  & \texttt{reverse\_word} & 0.00 & 0.07 & \textcolor{red}{\textbf{0.028}} & 0.05 & 0.47 & 0.185 \\
\midrule
Comp. & \texttt{sentiment\_flip} & 0.04 & 0.05 & \textcolor{red}{\textbf{0.068}} & 0.06 & 0.03 & 0.140 \\
\bottomrule
\end{tabular}
\end{table}

\begin{table}[!htbp]
\caption{Baselines and IID accuracy for Gemma-2-9B Base and Instruct. Format matches Table~\ref{tab:baselines_llama}.}
\label{tab:baselines_gemma}
\centering
\small
\begin{tabular}{@{}ll|ccc|ccc@{}}
\toprule
& & \multicolumn{3}{c|}{\textbf{Gemma-2 Base}} & \multicolumn{3}{c}{\textbf{Gemma-2 IT}} \\
\textbf{Cat.} & \textbf{Task} & Zero & 5-shot & IID & Zero & 5-shot & IID \\
\midrule
\multirow{3}{*}{Lex.}
  & \texttt{antonym} & 0.59 & 0.68 & 0.867 & 0.63 & 0.68 & 0.880 \\
  & \texttt{synonym} & 0.17 & 0.48 & 0.440 & 0.29 & 0.57 & 0.675 \\
  & \texttt{hypernym} & 0.21 & 0.92 & 0.902 & 0.14 & 0.88 & 0.897 \\
\midrule
\multirow{3}{*}{Fact.}
  & \texttt{country\_capital} & 0.42 & 1.00 & 0.930 & 0.53 & 1.00 & 0.982 \\
  & \texttt{english\_spanish} & 0.61 & 1.00 & 0.897 & 0.62 & 0.99 & 0.905 \\
  & \texttt{object\_color} & 0.09 & 0.12 & 0.522 & 0.06 & 0.19 & 0.575 \\
\midrule
\multirow{2}{*}{Morph.}
  & \texttt{past\_tense} & 0.56 & 1.00 & 0.948 & 0.57 & 1.00 & 0.982 \\
  & \texttt{plural} & 0.65 & 1.00 & 0.998 & 0.71 & 1.00 & 1.000 \\
\midrule
\multirow{3}{*}{Char.}
  & \texttt{capitalize} & 0.06 & 1.00 & 0.718 & 0.40 & 1.00 & 0.865 \\
  & \texttt{first\_letter} & 0.45 & 0.36 & 0.613 & 0.41 & 0.15 & 0.748 \\
  & \texttt{reverse\_word} & 0.04 & 0.28 & \textcolor{red}{\textbf{0.085}} & 0.34 & 0.26 & 0.393 \\
\midrule
Comp. & \texttt{sentiment\_flip} & 0.02 & 0.05 & 0.125 & 0.03 & 0.01 & 0.198 \\
\bottomrule
\end{tabular}
\end{table}

\begin{table}[!htbp]
\caption{Baselines and IID accuracy for Mistral-7B-v0.3 Base and Instruct. Mistral zero-shot accuracies are notably lower than Llama and Gemma for lexical tasks, reflecting weaker few-shot formatting sensitivity.}
\label{tab:baselines_mistral}
\centering
\small
\begin{tabular}{@{}ll|ccc|ccc@{}}
\toprule
& & \multicolumn{3}{c|}{\textbf{Mistral Base}} & \multicolumn{3}{c}{\textbf{Mistral Instruct}} \\
\textbf{Cat.} & \textbf{Task} & Zero & 5-shot & IID & Zero & 5-shot & IID \\
\midrule
\multirow{3}{*}{Lex.}
  & \texttt{antonym} & 0.02 & 0.68 & 0.740 & 0.02 & 0.66 & 0.780 \\
  & \texttt{synonym} & 0.01 & 0.16 & 0.138 & 0.09 & 0.18 & 0.367 \\
  & \texttt{hypernym} & 0.09 & 0.12 & 0.315 & 0.07 & 0.12 & 0.435 \\
\midrule
\multirow{3}{*}{Fact.}
  & \texttt{country\_capital} & 0.14 & 0.18 & 0.240 & 0.02 & 0.10 & 0.265 \\
  & \texttt{english\_spanish} & 0.59 & 0.76 & 0.818 & 0.60 & 0.77 & 0.797 \\
  & \texttt{object\_color} & 0.12 & 0.28 & 0.578 & 0.02 & 0.28 & 0.688 \\
\midrule
\multirow{2}{*}{Morph.}
  & \texttt{past\_tense} & 0.51 & 0.93 & 0.823 & 0.67 & 0.93 & 0.895 \\
  & \texttt{plural} & 0.58 & 0.85 & 0.940 & 0.83 & 0.85 & 0.983 \\
\midrule
\multirow{3}{*}{Char.}
  & \texttt{capitalize} & 0.01 & 0.08 & \textcolor{red}{\textbf{0.030}} & 0.03 & 0.08 & \textcolor{red}{\textbf{0.088}} \\
  & \texttt{first\_letter} & 0.23 & 0.43 & 0.573 & 0.11 & 0.30 & 0.723 \\
  & \texttt{reverse\_word} & 0.00 & 0.15 & \textcolor{red}{\textbf{0.030}} & 0.01 & 0.07 & \textcolor{red}{\textbf{0.037}} \\
\midrule
Comp. & \texttt{sentiment\_flip} & 0.03 & 0.03 & \textcolor{red}{\textbf{0.045}} & 0.00 & 0.00 & 0.095 \\
\bottomrule
\end{tabular}
\end{table}

\paragraph{Aggregate destructive cases.} Across all 6 models, the rate at which the mean steered accuracy drops below the zero-shot baseline---i.e., FV steering hurts the task in aggregate---is 0--1\%. As discussed in \S\ref{sec:iid}, this aggregate-task statistic is compatible with prior reports of substantial \emph{per-sample} anti-steerability~\citep{tan2024analysing, braun2025understanding}: positive effects can dominate the mean even when individual examples are pushed in the wrong direction. The aggregate result establishes that FV injection at the task-mean level is consistently additive or neutral; it does not contradict the existence of per-sample failures documented in prior work.

\FloatBarrier
\section{Per-Template IID Accuracy}
\label{app:per_template}

Per-template variance is substantial within several tasks. Table~\ref{tab:per_template_llama} shows per-template IID accuracy for Llama-3.1-8B Base. \texttt{antonym} ranges from 0.10 (T2) to 0.92 (T8); \texttt{country\_capital} from 0.22 (T2) to 0.88 (T4); \texttt{plural} achieves 1.00 on 6 of 8 templates. Per-template variance is one source of IID--OOD gap heterogeneity: a strong source template transfers well, a weak one transfers poorly.

\begin{table}[!htbp]
\caption{Per-template IID accuracy for Llama-3.1-8B Base. Each cell is the best accuracy across all $\alpha$ values and layers for the (task, template) pair.}
\label{tab:per_template_llama}
\centering
\small
\begin{tabular}{@{}l*{8}{c}|c@{}}
\toprule
\textbf{Task} & T1 & T2 & T3 & T4 & T5 & T6 & T7 & T8 & \textbf{Mean} \\
\midrule
\texttt{antonym}         & .90 & .10 & .60 & .88 & .76 & .86 & .88 & .92 & .738 \\
\texttt{synonym}         & .54 & .60 & .04 & .28 & .38 & .42 & .32 & .28 & .357 \\
\texttt{hypernym}        & .14 & .02 & .40 & .00 & .02 & .02 & .04 & .04 & .085 \\
\texttt{country\_capital}   & .86 & .22 & .34 & .88 & .58 & .76 & .84 & .84 & .665 \\
\texttt{english\_spanish}     & .94 & .92 & .56 & .90 & .72 & .80 & .90 & .92 & .833 \\
\texttt{object\_color}   & .34 & .48 & .28 & .32 & .30 & .40 & .38 & .36 & .357 \\
\texttt{past\_tense}     & 1.00 & .68 & .62 & 1.00 & .90 & .96 & .98 & 1.00 & .892 \\
\texttt{plural}          & 1.00 & .16 & .60 & 1.00 & 1.00 & 1.00 & 1.00 & 1.00 & .845 \\
\texttt{capitalize}      & .00 & .90 & .16 & .48 & .46 & .40 & .66 & .24 & .413 \\
\texttt{first\_letter}   & .44 & .32 & .46 & .34 & .30 & .28 & .36 & .36 & .357 \\
\texttt{reverse\_word}   & .12 & .00 & .00 & .00 & .06 & .00 & .02 & .02 & .028 \\
\texttt{sentiment\_flip}     & .14 & .06 & .06 & .08 & .04 & .04 & .08 & .04 & .068 \\
\bottomrule
\end{tabular}
\end{table}

\FloatBarrier
\section{Within-Style vs.\ Across-Style Transfer}
\label{app:style}

Table~\ref{tab:style} reports within-style and across-style mean OOD transfer accuracies for all 6 models. The tested null hypothesis is that within-style pairs (e.g., Natural$\to$Natural) achieve higher transfer than across-style pairs (e.g., Natural$\to$Symbolic). It is not rejected on any model: $p > 0.30$ in every case. This is a positive finding---a quantitative confirmation, with formal statistical tests, of the qualitative observation by \citet{todd2024function} that FVs are ``remarkably robust'' across surface contexts and the finding of \citet{braun2025understanding} that no prompt type consistently dominates: FVs capture an abstract, style-invariant computational instruction not tied to the surface form of the extraction template.

\begin{table}[!htbp]
\caption{Within-style vs.\ across-style mean OOD transfer accuracy. All 6 models show no significant style effect ($p > 0.30$).}
\label{tab:style}
\centering
\small
\begin{tabular}{@{}lcccc@{}}
\toprule
\textbf{Model} & \textbf{Within-style mean} & \textbf{Across-style mean} & $t$ & $p$ \\
\midrule
Gemma-2 Base & 0.613 & 0.644 & $-0.84$ & 0.40 \\
Gemma-2 IT & 0.723 & 0.735 & $-0.35$ & 0.72 \\
Llama-3.1 Base & 0.424 & 0.438 & $-0.36$ & 0.72 \\
Llama-3.1 IT & 0.663 & 0.663 & $+0.01$ & 0.99 \\
Mistral Base & 0.400 & 0.416 & $-0.42$ & 0.67 \\
Mistral IT & 0.456 & 0.478 & $-0.56$ & 0.58 \\
\bottomrule
\end{tabular}
\end{table}

\FloatBarrier
\section{Cross-Template Transfer per Model}
\label{app:transfer}

Table~\ref{tab:transfer_per_model} reports the source-IID-vs-OOD-transfer correlation across all 6 models, plus the IID--OOD gap aggregated over IID-gated tasks. The strong source-IID--transfer correlation ($r = 0.80$--$0.93$) confirms that cross-template ``failure'' is primarily a source-quality problem: weak source templates produce weak FVs, which transfer weakly to all targets; strong source templates transfer well across all 7 target templates.

\begin{table}[!htbp]
\caption{Source-template IID quality strongly predicts OOD transfer accuracy across all 6 models. ``Mean OOD gap'' is averaged over tasks passing the IID gate ($\tau = 0.10$). Negative gap indicates OOD is below IID; positive gap indicates the opposite (e.g., \texttt{plural} on Gemma where OOD slightly exceeds IID due to template-specific noise).}
\label{tab:transfer_per_model}
\centering
\small
\begin{tabular}{@{}lccc@{}}
\toprule
\textbf{Model} & $r$(\textbf{source IID}, \textbf{OOD}) & \textbf{Mean IID--OOD gap} & \textbf{Pairs analyzed} \\
\midrule
Llama-3.1 Base & $+0.87$ & $-0.043$ & 4{,}616 \\
Llama-3.1 Instruct & $+0.81$ & $-0.039$ & 5{,}352 \\
Gemma-2 Base & $+0.93$ & $-0.024$ & 6{,}099 \\
Gemma-2 Instruct & $+0.91$ & $-0.019$ & 5{,}352 \\
Mistral Base & $+0.80$ & $-0.058$ & 4{,}044 \\
Mistral Instruct & $+0.85$ & $-0.041$ & 4{,}417 \\
\bottomrule
\end{tabular}
\end{table}

\FloatBarrier
\section{Geometric Analysis Details}
\label{app:geometric}

Tables~\ref{tab:within_llama}, \ref{tab:within_gemma}, and \ref{tab:within_mistral} report per-task alignment--transfer correlations for all 6 models. The pattern in the main text---pooled $r$ near zero (or weakly negative on Mistral), with within-task $r$ scattered around zero---is consistent across families. Mistral IT is the most informative case: pooled $r = -0.187$, yet mean within-task $r = +0.139$, with \texttt{object\_color} driving a strong positive within-task correlation ($r = +0.589$). This is a textbook Simpson's paradox.

\begin{table}[!htbp]
\caption{Within-task alignment--transfer correlations (Llama-3.1-8B Base). Pooled $r = +0.013$ ($p = 0.16$, not significant). The hypernym within-task correlation ($r = -0.510$) is a floor-effect artifact: hypernym fails the IID gate on Llama Base (mean IID 0.085), so its within-task variation is dominated by noise.}
\label{tab:within_llama}
\centering
\small
\begin{tabular}{@{}llcc@{}}
\toprule
\textbf{Cat.} & \textbf{Task} & $r$ & $p$ \\
\midrule
Lex. & \texttt{antonym} & $+0.053$ & 0.116 \\
Lex. & \texttt{synonym} & $-0.008$ & 0.800 \\
Lex. & \texttt{hypernym} & $-0.510$ & $2.3\times10^{-60}$ \\
Fact. & \texttt{country\_capital} & $+0.050$ & 0.137 \\
Fact. & \texttt{english\_spanish} & $-0.118$ & $3.9\times10^{-4}$ \\
Fact. & \texttt{object\_color} & $+0.276$ & $4.2\times10^{-17}$ \\
Morph. & \texttt{past\_tense} & $+0.139$ & $3.0\times10^{-5}$ \\
Morph. & \texttt{plural} & $+0.248$ & $4.9\times10^{-14}$ \\
Char. & \texttt{capitalize} & $+0.284$ & $4.3\times10^{-18}$ \\
Char. & \texttt{first\_letter} & $-0.348$ & $6.8\times10^{-27}$ \\
Char. & \texttt{reverse\_word} & $-0.030$ & 0.373 \\
Comp. & \texttt{sentiment\_flip} & $+0.044$ & 0.185 \\
\midrule
\textbf{Pooled} &  & $+0.013$ & 0.163 \\
Mean within-task $r$ &  & $+0.007$ & --- \\
\bottomrule
\end{tabular}
\end{table}

\begin{table}[!htbp]
\caption{Within-task alignment--transfer correlations for Gemma-2-9B Base and Instruct.}
\label{tab:within_gemma}
\centering
\small
\begin{tabular}{@{}ll|cc|cc@{}}
\toprule
& & \multicolumn{2}{c|}{\textbf{Gemma Base}} & \multicolumn{2}{c}{\textbf{Gemma IT}} \\
\textbf{Cat.} & \textbf{Task} & $r$ & $p$ & $r$ & $p$ \\
\midrule
Lex. & \texttt{antonym} & $+0.142$ & $1.1\times10^{-6}$ & $+0.340$ & $3.0\times10^{-33}$ \\
Lex. & \texttt{synonym} & $+0.007$ & 0.82 & $-0.114$ & $8.3\times10^{-5}$ \\
Lex. & \texttt{hypernym} & $+0.040$ & 0.17 & $+0.009$ & 0.75 \\
Fact. & \texttt{country\_capital} & $-0.267$ & $1.2\times10^{-20}$ & $-0.036$ & 0.22 \\
Fact. & \texttt{english\_spanish} & $-0.081$ & $5.4\times10^{-3}$ & $+0.002$ & 0.95 \\
Fact. & \texttt{object\_color} & $-0.137$ & $2.4\times10^{-6}$ & $+0.263$ & $4.0\times10^{-20}$ \\
Morph. & \texttt{past\_tense} & $+0.117$ & $5.7\times10^{-5}$ & $+0.062$ & 0.034 \\
Morph. & \texttt{plural} & $-0.146$ & $4.8\times10^{-7}$ & $-0.035$ & 0.24 \\
Char. & \texttt{capitalize} & $+0.138$ & $2.1\times10^{-6}$ & $+0.278$ & $2.4\times10^{-22}$ \\
Char. & \texttt{first\_letter} & $-0.028$ & 0.33 & $+0.115$ & $7.6\times10^{-5}$ \\
Char. & \texttt{reverse\_word} & $-0.072$ & 0.014 & $+0.138$ & $2.0\times10^{-6}$ \\
Comp. & \texttt{sentiment\_flip} & $+0.075$ & 0.010 & $-0.029$ & 0.32 \\
\midrule
\textbf{Pooled} & & $+0.058$ & $3.9\times10^{-12}$ & $+0.126$ & $3.0\times10^{-51}$ \\
Mean within-task $r$ & & $-0.018$ & --- & $+0.083$ & --- \\
\bottomrule
\end{tabular}
\end{table}

\begin{table}[!htbp]
\caption{Within-task alignment--transfer correlations for Mistral-7B-v0.3. Pooled $r$ is negative for both, yet Mistral IT mean within-task $r = +0.139$---a genuine Simpson's paradox.}
\label{tab:within_mistral}
\centering
\small
\begin{tabular}{@{}ll|cc|cc@{}}
\toprule
& & \multicolumn{2}{c|}{\textbf{Mistral Base}} & \multicolumn{2}{c}{\textbf{Mistral IT}} \\
\textbf{Cat.} & \textbf{Task} & $r$ & $p$ & $r$ & $p$ \\
\midrule
Lex. & \texttt{antonym} & $+0.255$ & $9.3\times10^{-15}$ & $+0.281$ & $9.6\times10^{-18}$ \\
Lex. & \texttt{synonym} & $+0.067$ & 0.045 & $-0.023$ & 0.50 \\
Lex. & \texttt{hypernym} & $-0.461$ & $2.0\times10^{-48}$ & $+0.071$ & 0.034 \\
Fact. & \texttt{country\_capital} & $+0.017$ & 0.62 & $+0.280$ & $1.3\times10^{-17}$ \\
Fact. & \texttt{english\_spanish} & $-0.028$ & 0.41 & $+0.035$ & 0.29 \\
Fact. & \texttt{object\_color} & $+0.282$ & $8.1\times10^{-18}$ & $+0.589$ & $7.5\times10^{-85}$ \\
Morph. & \texttt{past\_tense} & $+0.080$ & 0.016 & $+0.181$ & $4.6\times10^{-8}$ \\
Morph. & \texttt{plural} & $+0.155$ & $3.0\times10^{-6}$ & $+0.067$ & 0.044 \\
Char. & \texttt{capitalize} & $+0.063$ & 0.060 & $+0.158$ & $1.9\times10^{-6}$ \\
Char. & \texttt{first\_letter} & $-0.290$ & $7.2\times10^{-19}$ & $-0.124$ & $2.1\times10^{-4}$ \\
Char. & \texttt{reverse\_word} & $-0.195$ & $3.7\times10^{-9}$ & $-0.148$ & $8.7\times10^{-6}$ \\
Comp. & \texttt{sentiment\_flip} & $-0.080$ & 0.017 & $+0.296$ & $1.6\times10^{-19}$ \\
\midrule
\textbf{Pooled} & & $-0.199$ & $1.2\times10^{-96}$ & $-0.187$ & $2.0\times10^{-85}$ \\
Mean within-task $r$ & & $-0.011$ & --- & $+0.139$ & --- \\
\bottomrule
\end{tabular}
\end{table}

\paragraph{Geometric dissociation cases.} Defining a dissociation as a cross-template pair with cosine alignment $> 0.80$ but transfer accuracy $< 0.40$ (a fixed-threshold scheme), Table~\ref{tab:dissociation} reports per-model dissociation counts. We additionally verified that the qualitative conclusion---steering accuracy exceeds geometric similarity as a predictor of transfer---is invariant to the threshold scheme. Re-running the analysis with data-derived percentile thresholds (top-20\% cosine vs.\ bottom-50\% accuracy) yields the same population-level conclusions on all 6 models, though absolute counts differ.

\begin{table}[!htbp]
\caption{Geometric dissociation: cross-template pairs with cosine $> 0.80$ but transfer accuracy $< 0.40$. ``IID-Viable'' = dissociations where the source template passes the IID gate, isolating geometry--behavior dissociation from extraction failure.}
\label{tab:dissociation}
\centering
\small
\begin{tabular}{@{}lcccc@{}}
\toprule
\textbf{Model} & \textbf{High-Align Pairs} & $N_{\text{dissociation}}$ & \textbf{Rate} & \textbf{IID-Viable} \\
\midrule
Llama Base & 4{,}470 & 6{,}657 & 0.745 & 4{,}616 \\
Llama IT & 3{,}197 & 2{,}705 & 0.423 & 2{,}345 \\
Gemma Base & 6{,}351 & 6{,}669 & 0.525 & 6{,}099 \\
Gemma IT & 5{,}535 & 3{,}860 & 0.349 & 3{,}699 \\
Mistral Base & 4{,}600 & 7{,}139 & 0.776 & 4{,}044 \\
Mistral IT & 4{,}212 & 6{,}140 & 0.729 & 4{,}417 \\
\bottomrule
\end{tabular}
\end{table}

\FloatBarrier
\section{Permutation Tests for Excess Dissociation}
\label{app:permutation}

For each task on each model, we test whether the observed dissociation rate exceeds chance using 1{,}000 permutations of the alignment--transfer pairing. Per-task $p$-values are Bonferroni-corrected for the family of 12 within-task tests ($\alpha_{\text{corrected}} = 0.0042$). Significant excess dissociation is task- and model-specific: Gemma Base shows significant excess for 6 of 12 tasks (\texttt{country\_capital}, \texttt{english\_spanish}, \texttt{hypernym}, \texttt{past\_tense}, \texttt{plural}, \texttt{synonym}); Gemma IT for 3 of 12 (\texttt{english\_spanish}, \texttt{reverse\_word}, \texttt{synonym}); Llama Base for 1 of 12 (\texttt{country\_capital}); Mistral Base for 1 of 12 (\texttt{first\_letter}); Llama IT and Mistral IT show no tasks significant after correction. The pattern reflects task-specific geometric structure rather than a uniform artifact.

\FloatBarrier
\section{Universal Template Vector (UTV) Analysis}
\label{app:utv}

We apply PCA to the 8 template FVs per task at the best steering layer and report the fraction of variance explained by PC1. High PC1 indicates templates converge on a shared FV direction; low PC1 indicates templates extract different directions.

\begin{table}[!htbp]
\caption{PC1 variance explained (range across 6 models) for each task. Tasks sorted by mean PC1.}
\label{tab:utv}
\centering
\small
\begin{tabular}{@{}llcc@{}}
\toprule
\textbf{Cat.} & \textbf{Task} & \textbf{PC1 Range} & \textbf{Mean PC1} \\
\midrule
Char. & \texttt{reverse\_word} & 0.55--0.97 & 0.80 \\
Lex. & \texttt{antonym} & 0.52--0.92 & 0.72 \\
Morph. & \texttt{plural} & 0.44--0.80 & 0.63 \\
Char. & \texttt{capitalize} & 0.42--0.84 & 0.60 \\
Morph. & \texttt{past\_tense} & 0.38--0.73 & 0.55 \\
Fact. & \texttt{english\_spanish} & 0.35--0.68 & 0.50 \\
Fact. & \texttt{country\_capital} & 0.33--0.65 & 0.48 \\
Lex. & \texttt{synonym} & 0.37--0.63 & 0.47 \\
Comp. & \texttt{sentiment\_flip} & 0.30--0.62 & 0.44 \\
Lex. & \texttt{hypernym} & 0.32--0.62 & 0.43 \\
Char. & \texttt{first\_letter} & 0.30--0.58 & 0.42 \\
Fact. & \texttt{object\_color} & 0.36--0.55 & 0.42 \\
\bottomrule
\end{tabular}
\end{table}

Strikingly, high PC1 does not predict high transfer: \texttt{reverse\_word} has the highest template convergence (PC1 = 0.80) yet near-zero IID accuracy on most models (Table~\ref{tab:baselines_llama}--\ref{tab:baselines_mistral}), while \texttt{synonym} and \texttt{object\_color} have low PC1 (0.42--0.47) yet moderate-to-good steering. This confirms that the geometry of template FVs (even their shared direction structure) does not determine functional effectiveness---reinforcing the §\ref{sec:geometric} hierarchical-regression conclusion at a different level of analysis.

\FloatBarrier
\section{FV Norm Analysis}
\label{app:norms}

Table~\ref{tab:norms} reports mean FV L2 norms and FV-norm--transfer correlations across all cross-template pairs. Gemma FV norms (277--283) are 15--30$\times$ larger than Mistral's (9.7--12.6) and 12--17$\times$ larger than Llama's (16.4--20.5), likely reflecting architectural and normalization differences (Gemma-2's $d_{\text{model}} = 3584$ with 42 layers vs.\ Llama and Mistral's $d_{\text{model}} = 4096$ with 32 layers, plus differing residual-stream normalization conventions).

\begin{table}[!htbp]
\caption{FV L2 norms (mean across all extracted FVs at the best steering layer) and norm--transfer Pearson correlations across all cross-template pairs ($n \approx$ 10{,}752--14{,}112 pairs per model).}
\label{tab:norms}
\centering
\small
\begin{tabular}{@{}lccc@{}}
\toprule
\textbf{Model} & \textbf{Mean Norm} & $r(\text{norm, transfer})$ & $p$ \\
\midrule
Llama-3.1 Base & 20.50 & $-0.089$ & $1.9\times10^{-20}$ \\
Llama-3.1 IT & 16.39 & $-0.077$ & $1.9\times10^{-15}$ \\
Gemma-2 Base & 283.12 & $-0.034$ & $6.5\times10^{-5}$ \\
Gemma-2 IT & 277.09 & $-0.046$ & $3.6\times10^{-8}$ \\
Mistral Base & 9.71 & $-0.013$ & 0.18 \\
Mistral IT & 12.64 & $+0.132$ & $5.8\times10^{-43}$ \\
\bottomrule
\end{tabular}
\end{table}

The norm--transfer correlation is weakly negative or near zero for 5 of 6 models; only Mistral IT shows a positive relationship ($r = +0.132$). This is the third independent line of evidence (after post-steering logit-lens deltas and patching recovery) for the dual-mechanism hypothesis: Mistral's effect on transfer is genuinely magnitude-driven, while Llama's and Gemma's are not. Notably, despite Gemma's norms being 15--30$\times$ larger than Mistral's, Gemma's per-layer post-steering deltas are smaller (\S\ref{app:post_steering}). Magnitude is not what drives readability change; alignment with $W_U$ is. This partially refutes a ``more push = more change'' hypothesis and reinforces the modulatory interpretation of Llama and Gemma FV mechanisms.

\FloatBarrier
\section{Instruction-Tuning Effect}
\label{app:instruct}

Table~\ref{tab:instruct} reports the per-task instruction-tuning effect ($\Delta = \mathrm{IT} - \mathrm{Base}$). Instruction tuning consistently improves FV steering accuracy: mean $\Delta$ is $+$0.21 (Llama), $+$0.09 (Gemma), $+$0.07 (Mistral). The dissociation pattern is preserved across the IT--Base axis, indicating instruction tuning improves IID efficacy without changing the qualitative conclusion that FV steering operates outside the logit-lens-decodable subspace.

\begin{table}[!htbp]
\caption{Instruction-tuning effect on IID steering accuracy ($\Delta = $ Instruct $-$ Base).}
\label{tab:instruct}
\centering
\small
\begin{tabular}{@{}lccc@{}}
\toprule
\textbf{Task} & \textbf{Gemma $\Delta$} & \textbf{Llama $\Delta$} & \textbf{Mistral $\Delta$} \\
\midrule
\texttt{antonym} & $+0.013$ & $+0.132$ & $+0.040$ \\
\texttt{synonym} & $+0.235$ & $+0.268$ & $+0.229$ \\
\texttt{hypernym} & $-0.005$ & $+0.327$ & $+0.120$ \\
\texttt{country\_capital} & $+0.052$ & $+0.255$ & $+0.025$ \\
\texttt{english\_spanish} & $+0.008$ & $+0.049$ & $-0.021$ \\
\texttt{object\_color} & $+0.053$ & $+0.363$ & $+0.110$ \\
\texttt{past\_tense} & $+0.034$ & $+0.051$ & $+0.072$ \\
\texttt{plural} & $+0.002$ & $+0.138$ & $+0.043$ \\
\texttt{capitalize} & $+0.147$ & $+0.312$ & $+0.058$ \\
\texttt{first\_letter} & $+0.135$ & $+0.420$ & $+0.150$ \\
\texttt{reverse\_word} & $+0.308$ & $+0.157$ & $+0.007$ \\
\texttt{sentiment\_flip} & $+0.073$ & $+0.072$ & $+0.050$ \\
\midrule
\textbf{Mean} & $+0.088$ & $+0.212$ & $+0.074$ \\
\bottomrule
\end{tabular}
\end{table}

\FloatBarrier
\section{Pre-Registered Predictions and Outcomes}
\label{app:predictions}

\begin{table}[!htbp]
\caption{Pre-registered predictions vs.\ observed grand mean across 6 models. \checkmark{} = confirmed within range; $\sim$ = partially confirmed (slightly above); $\times$ = falsified. Predictions were committed before any model runs.}
\label{tab:predictions}
\centering
\small
\begin{tabular}{@{}llccl@{}}
\toprule
\textbf{Task} & \textbf{Predicted} & \textbf{Observed} & \textbf{Status} & \textbf{Notes} \\
\midrule
\texttt{antonym} & 0.45--0.60 & 0.813 & $\times$ & Exceeded range \\
\texttt{synonym} & 0.20--0.40 & 0.434 & $\sim$ & Slightly above \\
\texttt{hypernym} & 0.25--0.45 & 0.508 & $\sim$ & Strongly model-dependent \\
\texttt{country\_capital} & 0.40--0.65 & 0.667 & $\sim$ & Slightly above \\
\texttt{english\_spanish} & 0.25--0.50 & 0.856 & $\times$ & Greatly exceeded \\
\texttt{object\_color} & 0.30--0.55 & 0.573 & $\sim$ & Slightly above \\
\texttt{past\_tense} & 0.35--0.55 & 0.914 & $\times$ & Greatly exceeded \\
\texttt{plural} & 0.30--0.50 & 0.958 & $\times$ & Greatly exceeded \\
\texttt{capitalize} & 0.00--0.05 & 0.473 & $\times$ & Falsified neg.\ control \\
\texttt{first\_letter} & 0.05--0.20 & 0.632 & $\times$ & Falsified neg.\ control \\
\texttt{reverse\_word} & 0.00--0.08 & 0.126 & $\sim$ & Near predicted \\
\texttt{sentiment\_flip} & 0.00--0.05 & 0.112 & $\sim$ & Near predicted \\
\bottomrule
\end{tabular}
\end{table}

The prediction grid was conservative: 4 of 12 tasks were greatly exceeded (\texttt{antonym}, \texttt{english\_spanish}, \texttt{past\_tense}, \texttt{plural}), and 2 of 3 character ``negative controls'' (\texttt{capitalize}, \texttt{first\_letter}) were strongly falsified. Only \texttt{reverse\_word} and \texttt{sentiment\_flip} behaved as predicted. The honest assessment of these failures motivated the IID-gating definition (§\ref{def:iid_gate}): rather than retroactively redefine task difficulty, we use a single threshold to admit tasks into causal analysis.

\FloatBarrier
\section{Tuned-Lens Statistics}
\label{app:tuned_lens}

\paragraph{Training procedure.} Per-layer translators are trained independently for each layer using AdamW (learning rate $10^{-3}$, weight decay 0.01), early stopping with patience 15 epochs and improvement threshold $10^{-7}$, and a maximum of 200 epochs. Training data are pooled zero-shot residual-stream activations across all tasks and templates (12 tasks $\times$ 8 templates $\times$ ${\sim}85$ examples $\approx 8{,}176$ training points per layer); the target for every training point is the final-layer residual stream at the same example, regardless of task---translators are task-agnostic by construction. We use an 80\%/20\% train/validation split with seeded permutation and best-state restoration: if no epoch improves over the identity translator on validation MSE, the translator resets to identity.

\paragraph{Conservative design.} The diagonal constraint captures per-feature scaling and offset (the dominant source of representational dialect~\citep{belrose2023eliciting}) without introducing the capacity for complex learned classification. Compared to low-rank (rank-16, rank-64) tuned-lens variants, the diagonal lens is a more \emph{conservative} test: it cannot rotate the representational frame, so any failure to recover decodability is a stronger negative result than an equivalent failure under a rotation-capable translator. Off-diagonal rotations and nonlinear decoders are concrete next steps; if a 2-layer MLP decoder trained on intermediate activations to predict task outputs were to find the answer at early layers (where FVs intervene), the ``computational instruction'' interpretation would narrow to ``nonlinearly encoded instruction''---a strictly weaker claim. Our diagonal constraint, by contrast, leaves room for that future test to refute or refine our finding.

\paragraph{Paired tests: logit lens vs.\ tuned lens.} Across all task$\times$template$\times$layer observations, paired $t$-tests (Table~\ref{tab:tl_paired}) show that the tuned lens slightly \emph{degrades} top-10 readability on 4 of 6 models (Llama Base, Llama Instruct, Mistral Base, Mistral Instruct), is statistically indistinguishable on Gemma Base, and slightly degrades Gemma Instruct. Effect sizes are small (Cohen's $|d| \leq 0.27$). This is consistent with the dialect-correction--readability anti-correlation reported in §\ref{sec:tuned_lens_validation}: layers requiring the largest MSE-driven corrections show the greatest readability degradation, because the MSE objective optimizes for reconstructing dominant variance directions of the final residual stream---directions dominated by syntactic and positional information that interfere with the fragile task-answer signal.

\begin{table}[!htbp]
\caption{Paired $t$-tests: logit lens vs.\ tuned lens top-10 readability across all task$\times$template$\times$layer observations per model. The tuned lens slightly degrades readability on 4 of 6 models; effect sizes are small.}
\label{tab:tl_paired}
\centering
\small
\begin{tabular}{@{}lrcrrrc@{}}
\toprule
\textbf{Model} & $N$ & \textbf{LL Mean} & \textbf{TL Mean} & $\Delta$ & $t$ & \textbf{Cohen's} $d$ \\
\midrule
Llama-3.1 Base & 1{,}536 & 0.026 & 0.022 & $-0.004$ & $-9.19^{***}$ & $-0.23$ \\
Llama-3.1 Instruct & 1{,}536 & 0.044 & 0.036 & $-0.008$ & $-9.84^{***}$ & $-0.25$ \\
Mistral Base & 1{,}536 & 0.057 & 0.048 & $-0.009$ & $-10.50^{***}$ & $-0.27$ \\
Mistral Instruct & 1{,}536 & 0.055 & 0.050 & $-0.005$ & $-8.83^{***}$ & $-0.23$ \\
Gemma-2 Base & 2{,}016 & 0.022 & 0.022 & $+0.000$ & $+1.49$ & $+0.03$ \\
Gemma-2 Instruct & 2{,}016 & 0.039 & 0.038 & $-0.001$ & $-2.87^{**}$ & $-0.06$ \\
\bottomrule
\multicolumn{7}{@{}l}{\footnotesize $^{***}p < 10^{-5}$, $^{**}p < 0.01$.} \\
\end{tabular}
\end{table}

\paragraph{FV vocabulary projection coherence.} Table~\ref{tab:tl_fv} compares the correct-output fraction of FV vocabulary projections under the logit lens vs.\ the tuned lens. Both decoders yield negligible coherence ($<$6\%) on every model. Hypothesis ``H4'' (FVs encode answer information in a rotated dialect that the tuned lens can decode) is not supported on any model; the largest improvement is Mistral Base ($0.034 \to 0.053$), which still leaves $>$94\% of projected tokens unrelated to the correct output.

\begin{table}[!htbp]
\caption{FV vocabulary projection coherence: logit lens vs.\ tuned lens. ``Correct Frac'' = fraction of top-50 projected tokens matching any correct task output. Both decoders yield negligible coherence ($<$6\%); the rotated-dialect hypothesis (``H4?'') is not supported on any model.}
\label{tab:tl_fv}
\centering
\small
\begin{tabular}{@{}lcccrcc@{}}
\toprule
\textbf{Model} & \textbf{LL Frac} & \textbf{TL Frac} & $\Delta$ & $N_{\text{impr}}$ & $N_{\text{degr}}$ & \textbf{H4?} \\
\midrule
Llama-3.1 Base & 0.004 & 0.002 & $-0.001$ & 0 & 6 & No \\
Llama-3.1 Instruct & 0.003 & 0.012 & $+0.009$ & 24 & 3 & No \\
Mistral Base & 0.034 & 0.053 & $+0.019$ & 60 & 4 & No \\
Mistral Instruct & 0.035 & 0.037 & $+0.002$ & 15 & 13 & No \\
Gemma-2 Base & 0.017 & 0.018 & $+0.001$ & 4 & 2 & No \\
Gemma-2 Instruct & 0.014 & 0.014 & $+0.000$ & 2 & 1 & No \\
\bottomrule
\end{tabular}
\end{table}

\paragraph{Dialect-correction magnitude.} Table~\ref{tab:tl_dialect} reports the mean and maximum percentage reduction in MSE over the identity translator across trained layers per model, plus the Pearson correlation between layer depth and improvement percentage. Later layers universally require larger corrections ($r(\text{depth}, \text{improvement}) \geq +0.61$ on every model)---the ``dialect'' grows over depth, consistent with the residual stream gradually rotating away from the input embedding frame and toward the unembedding frame.

\begin{table}[!htbp]
\caption{Dialect correction magnitude by model family. ``Mean Improv.\ (\%)'' = mean percentage reduction in MSE over identity translator across trained layers. ``$r$(depth)'' = Pearson correlation between layer depth and improvement percentage. Later layers universally require larger corrections.}
\label{tab:tl_dialect}
\centering
\small
\begin{tabular}{@{}lcccc@{}}
\toprule
\textbf{Model} & \textbf{Mean Improv.\ (\%)} & \textbf{Max Improv.\ (\%)} & $r$(depth) & $p$ \\
\midrule
Llama-3.1 Base & 11.4 & 16.5 & $+0.898$ & $<10^{-4}$ \\
Llama-3.1 Instruct & 10.2 & 15.1 & $+0.867$ & $<10^{-4}$ \\
Mistral Base & 23.3 & 31.4 & $+0.892$ & $<10^{-4}$ \\
Mistral Instruct & 26.5 & 35.8 & $+0.912$ & $<10^{-4}$ \\
Gemma-2 Base & 3.4 & 14.1 & $+0.752$ & $0.0001$ \\
Gemma-2 Instruct & 2.3 & 19.7 & $+0.611$ & $0.004$ \\
\bottomrule
\end{tabular}
\end{table}

\paragraph{Expanded $2\times 3$ matrix.} The full expanded steerability--decodability matrix (Table~\ref{tab:2x3}) classifies each task$\times$model instance by logit-lens and tuned-lens top-10 accuracy crossing $\tau = 0.10$. The ``tuned lens only'' row---populated if dialect correction reveals previously hidden decodability---contains exactly 1 entry (the borderline \texttt{first\_letter}/Llama IT case). The binary classification of tasks as decodable vs.\ not decodable is therefore effectively decoder-invariant for this class of translators.

\begin{table}[!htbp]
\caption{Expanded $2\times 3$ steerability--decodability matrix aggregated across 6 models (72 task$\times$model pairs). A decoder ``succeeds'' if best top-10 accuracy exceeds 0.10 at any layer. The ``tuned lens only'' row is nearly empty, confirming the original $2\times 2$ classification is robust.}
\label{tab:2x3}
\centering
\small
\begin{tabular}{@{}l|cc|c@{}}
\toprule
& \textbf{Steerable} & \textbf{Not Steerable} & \textbf{Total} \\
\midrule
\textbf{Logit lens readable} & 54 (75.0\%) & 2 (2.8\%) & 56 \\
\textbf{Tuned lens only}    & 1 (1.4\%) & 0 (0.0\%) & 1 \\
\textbf{Neither readable}   & 13 (18.1\%) & 2 (2.8\%) & 15 \\
\midrule
\textbf{Total} & 68 & 4 & 72 \\
\bottomrule
\end{tabular}
\end{table}

\FloatBarrier
\section{Nonlinear-Decoder Follow-up: 2-Layer MLP Probe}
\label{app:mlp}

This appendix expands \S\ref{sec:mlp_decoder} of the main paper. The follow-up tests whether the \emph{steerable but not decodable} (SAND) cells survive a stronger decoder than the parameter-free logit lens or the diagonal tuned lens---namely a 2-layer MLP probe with a Hewitt \& Liang control task. The motivation is the obvious reviewer question: \emph{can a generic-capacity nonlinear decoder find the answer at the early layers where FVs intervene?} The result, summarized in the main body, decomposes the 10 logit-lens SAND cells into 5 closed by nonlinear decoding and 5 invisible to every decoder we tried, with the Hewitt \& Liang control catching 6 false-positive label-shortcut cases.

\subsection{Probe Architecture and Training}
\label{app:mlp:method}

For each (model, task, layer) triple we train a 2-layer feed-forward probe on pooled zero-shot residual-stream activations:
\begin{equation}
\hat{\mathbf{p}}(\mathbf{h}) \;=\; \mathrm{softmax}\!\left(W_2 \cdot \mathrm{Dropout}(\mathrm{GELU}(W_1 \cdot \mathrm{LayerNorm}(\mathbf{h}) + \mathbf{b}_1)) + \mathbf{b}_2\right),
\label{eq:mlp_probe}
\end{equation}
with hidden dimension $1024$, dropout $0.1$, and output dimension equal to the model's full vocabulary size $|V|$ (so the metric is comparable to the logit lens). Probes are trained for 30 epochs with full-batch AdamW (lr $10^{-3}$, weight decay $10^{-4}$, cosine schedule with 10\% warmup).

\paragraph{Split by unique input.} Following \citet{hewitt2019designing}, we split each task's input--output dataset 80/20 \emph{by unique input}, not by example. Because each input appears 8 times across the 8 templates, splitting by input rather than by row guarantees that the probe must generalize from training inputs to inputs it has \emph{never seen} (in any template). This is the strongest within-task generalization test compatible with our pooled-activation design and rules out a probe ``cheating'' by memorizing a template-specific surface form.

\paragraph{Hewitt \& Liang control task.} For every probe we train an \emph{identical} second probe on the same activations against a deterministic random shuffle of the input$\to$label mapping (seed $1234$): each input is reassigned a label uniformly at random from the task's label set, and that assignment is fixed across all 8 templates so that cross-template generalization remains intact---only the input$\to$label pairing is broken. Selectivity = $\text{(real probe top-10)} - \text{(control probe top-10)}$. The control captures raw decoder capacity (label-marginal memorization, vocabulary shortcuts, signal in features that correlate with labels but not with input). High real top-10 with low selectivity means the probe memorized; high selectivity means the probe found genuine input-conditional structure.

\paragraph{Decodability rule.} A (model, task) cell is declared MLP-decodable iff the best-layer real top-10 $\geq \tau = 0.10$ (matching the main paper's readability threshold) \emph{and} the corresponding selectivity $\geq \tau/2 = 0.05$. The selectivity gate is critical (\S\ref{app:mlp:selgate}): without it, 6 \texttt{object\_color} cells across all 6 models would have been falsely declared decodable.

\paragraph{Layer coverage.} We probe every other layer: 16 layers for Llama and Mistral (32 blocks), 21 layers for Gemma-2 (42 blocks). With 2 conditions (real/control), 12 tasks, and 6 models, the experiment trains 1{,}272 probes total ($\sim$50~min wall-clock on H200; full runtime budget in \S\ref{app:mlp:compute}).

\subsection{Per-Model Aggregate Statistics}
\label{app:mlp:per_model}

Table~\ref{tab:mlp_per_model} reports per-model aggregate statistics: number of probes trained, mean real and control top-10 across all (task, layer) pairs, mean selectivity, and the SAND-decomposition counts.

\begin{table}[!htbp]
\caption{Per-model MLP-probe summary. ``$n_{\text{probes}}$'' is the number of (task, layer) pairs in each condition (Llama/Mistral: 12 tasks $\times$ 16 layers $=$ 192; Gemma-2: 12 $\times$ 21 $=$ 252). ``Mean sel.'' is averaged across all probes per model (most are uninformative noise around zero, so this aggregate sits near zero by construction; the headline cases drive the per-task selectivities in Table~\ref{tab:mlp_per_task}). ``SAND'' columns count the SAND cells under the logit lens, the subset closed by the MLP, and the subset that persists.}
\label{tab:mlp_per_model}
\centering
\small
\begin{tabular}{@{}lrrrrrrr@{}}
\toprule
\textbf{Model} & $n_{\text{probes}}$ & \textbf{Mean real} & \textbf{Mean ctrl} & \textbf{Mean sel.} & \textbf{SAND} & \textbf{Closed} & \textbf{Persist} \\
\midrule
Llama-3.1 Base     & 192 & 0.179 & 0.180 & $-0.001$ & 4 & 2 & 2 \\
Llama-3.1 Instruct & 192 & 0.239 & 0.191 & $+0.048$ & 1 & 1 & 0 \\
Gemma-2 Base       & 252 & 0.226 & 0.160 & $+0.066$ & 0 & 0 & 0 \\
Gemma-2 Instruct   & 252 & 0.231 & 0.157 & $+0.074$ & 2 & 1 & 1 \\
Mistral Base       & 192 & 0.207 & 0.187 & $+0.021$ & 0 & 0 & 0 \\
Mistral Instruct   & 192 & 0.197 & 0.184 & $+0.012$ & 3 & 1 & 2 \\
\midrule
\textbf{Total}     & 1{,}272 & --- & --- & --- & \textbf{10} & \textbf{5} & \textbf{5} \\
\bottomrule
\end{tabular}
\end{table}

Several cross-model observations follow. (i) \textbf{Gemma-2 Base has zero SAND cells under the logit lens}; its instruct variant introduces 2 (\texttt{first\_letter}, \texttt{object\_color}). The base$\to$instruct fine-tune appears to displace part of the FV's representation off the unembedding axis on these two tasks. (ii) \textbf{Mean selectivity is positive for 5 of 6 models}; Llama Base sits at $-0.001$ (essentially noise). The high-selectivity per-task cases drive the headline finding---most (task, layer) pairs are uninformative, so the bulk of the 1{,}272 probes contribute random performance for both real and control. (iii) \textbf{Instruction-tuning changes mean selectivity differently per family}: Llama gains $+0.049$, Gemma gains $+0.008$ (already saturated at base), Mistral loses $0.008$. Llama is the model where the instruct fine-tune most dramatically reorganizes representations into probe-readable form.

\subsection{Per-Task, Per-Model Headline Numbers}
\label{app:mlp:per_task_per_model}

Table~\ref{tab:mlp_per_task} reports, for every (model, task) cell, the logit-lens best top-10, the MLP probe best real top-10, the MLP control top-10 at the same layer, the resulting selectivity, and the FV IID accuracy. Cells in the lower-left of the steerability--decodability matrix (steerable, not logit-lens decodable) are highlighted by their selectivity sign and value: positive$\geq$0.05 = SAND closed; negative or $<0.05$ = SAND persists.

\begin{small}
\begin{longtable}{@{}lllrrrrrr@{}}
\caption{Full per-(model, task) numerical summary of the MLP probe results. ``LL'' = logit lens best top-10; ``MLP real'' = MLP probe best top-10; ``MLP ctrl'' = control top-10 at the same layer; ``Sel'' = selectivity; ``FV IID'' = FV steering accuracy (mean across templates at best $\alpha$, layer); ``L*'' = best layer / total layers. Bold rows are the 10 logit-lens SAND cells (LL $<$ 0.10 \textbf{and} FV IID $\geq$ 0.10). $^{\dagger}$Closed by MLP probe. $^{\ddagger}$Persists under MLP probe.}
\label{tab:mlp_per_task}\\
\toprule
\textbf{Model} & \textbf{Task} & & \textbf{LL} & \textbf{MLP real} & \textbf{MLP ctrl} & \textbf{Sel} & \textbf{FV IID} & \textbf{L*} \\
\midrule
\endfirsthead
\toprule
\textbf{Model} & \textbf{Task} & & \textbf{LL} & \textbf{MLP real} & \textbf{MLP ctrl} & \textbf{Sel} & \textbf{FV IID} & \textbf{L*} \\
\midrule
\endhead
\midrule
\multicolumn{9}{r}{\textit{Continued on next page}} \\
\endfoot
\bottomrule
\endlastfoot
\multirow{12}{*}{Llama Base}
 & \texttt{antonym}         &              & 0.621 & 0.079 & 0.020 & $+0.06$ & 0.92 & 28/32 \\
 & \texttt{capitalize}      &              & 0.119 & 0.044 & 0.000 & $+0.04$ & 0.90 & 8/32  \\
 & \texttt{country\_capital}& $^{\dagger}$ & \textbf{0.056} & \textbf{0.396} & 0.125 & $+0.27$ & 0.88 & 2/32  \\
 & \texttt{english\_spanish}&              & 0.420 & 0.229 & 0.111 & $+0.12$ & 0.94 & 26/32 \\
 & \texttt{first\_letter}   & $^{\ddagger}$& \textbf{0.000} & \textbf{0.331} & 0.309 & $+0.02$ & 0.46 & 28/32 \\
 & \texttt{hypernym}        & $^{\dagger}$ & \textbf{0.023} & \textbf{0.441} & 0.184 & $+0.26$ & 0.40 & 6/32  \\
 & \texttt{object\_color}   & $^{\ddagger}$& \textbf{0.024} & \textbf{0.824} & 0.882 & $-0.06$ & 0.48 & 4/32  \\
 & \texttt{past\_tense}     &              & 0.678 & 0.056 & 0.049 & $+0.01$ & 1.00 & 16/32 \\
 & \texttt{plural}          &              & 0.700 & 0.056 & 0.146 & $-0.09$ & 1.00 & 20/32 \\
 & \texttt{reverse\_word}   &              & 0.113 & 0.125 & 0.172 & $-0.05$ & 0.12 & 22/32 \\
 & \texttt{sentiment\_flip} &              & 1.000 & 0.333 & 0.250 & $+0.08$ & 0.14 & 2/32  \\
 & \texttt{synonym}         &              & 0.193 & 0.090 & 0.028 & $+0.06$ & 0.60 & 28/32 \\
\midrule
\multirow{12}{*}{Llama IT}
 & \texttt{antonym}         &              & 0.579 & 0.092 & 0.039 & $+0.05$ & 0.92 & 26/32 \\
 & \texttt{capitalize}      &              & 0.345 & 0.044 & 0.000 & $+0.04$ & 1.00 & 4/32  \\
 & \texttt{country\_capital}&              & 0.333 & 0.306 & 0.042 & $+0.26$ & 0.98 & 14/32 \\
 & \texttt{english\_spanish}&              & 0.477 & 0.250 & 0.083 & $+0.17$ & 0.96 & 32/32 \\
 & \texttt{first\_letter}   & $^{\dagger}$ & \textbf{0.047} & \textbf{0.493} & 0.309 & $+0.18$ & 0.96 & 32/32 \\
 & \texttt{hypernym}        &              & 0.395 & 0.831 & 0.206 & $+0.62$ & 0.84 & 18/32 \\
 & \texttt{object\_color}   &              & 0.388 & 0.882 & 0.956 & $-0.07$ & 0.84 & 24/32 \\
 & \texttt{past\_tense}     &              & 0.633 & 0.056 & 0.035 & $+0.02$ & 1.00 & 28/32 \\
 & \texttt{plural}          &              & 0.678 & 0.069 & 0.153 & $-0.08$ & 1.00 & 26/32 \\
 & \texttt{reverse\_word}   &              & 0.175 & 0.102 & 0.094 & $+0.01$ & 0.48 & 24/32 \\
 & \texttt{sentiment\_flip} &              & 0.917 & 0.313 & 0.302 & $+0.01$ & 0.22 & 20/32 \\
 & \texttt{synonym}         &              & 0.216 & 0.062 & 0.014 & $+0.05$ & 0.70 & 4/32  \\
\midrule
\multirow{12}{*}{Gemma-2 Base}
 & \texttt{antonym}         &              & 0.611 & 0.158 & 0.046 & $+0.11$ & 0.90 & 32/42 \\
 & \texttt{capitalize}      &              & 0.286 & 0.059 & 0.000 & $+0.06$ & 0.96 & 6/42  \\
 & \texttt{country\_capital}&              & 0.933 & 0.000 & 0.056 & $-0.06$ & 0.98 & 2/42  \\
 & \texttt{english\_spanish}&              & 0.432 & 0.000 & 0.000 & $\phantom{+}0.00$ & 0.94 & 2/42 \\
 & \texttt{first\_letter}   &              & 0.105 & 0.860 & 0.397 & $+0.46$ & 1.00 & 40/42 \\
 & \texttt{hypernym}        &              & 0.558 & 0.934 & 0.199 & $+0.74$ & 0.98 & 42/42 \\
 & \texttt{object\_color}   &              & 0.729 & 0.882 & 0.963 & $-0.08$ & 0.84 & 34/42 \\
 & \texttt{past\_tense}     &              & 0.667 & 0.000 & 0.000 & $\phantom{+}0.00$ & 1.00 & 2/42 \\
 & \texttt{plural}          &              & 0.800 & 0.000 & 0.000 & $\phantom{+}0.00$ & 1.00 & 2/42 \\
 & \texttt{reverse\_word}   &              & 0.163 & 0.422 & 0.109 & $+0.31$ & 0.34 & 40/42 \\
 & \texttt{sentiment\_flip} &              & 1.000 & 0.250 & 0.333 & $-0.08$ & 0.22 & 18/42 \\
 & \texttt{synonym}         &              & 0.136 & 0.007 & 0.000 & $+0.01$ & 0.60 & 42/42 \\
\midrule
\multirow{12}{*}{Gemma-2 IT}
 & \texttt{antonym}         &              & 0.768 & 0.158 & 0.039 & $+0.12$ & 0.96 & 30/42 \\
 & \texttt{capitalize}      &              & 0.774 & 0.051 & 0.000 & $+0.05$ & 1.00 & 18/42 \\
 & \texttt{country\_capital}&              & 0.822 & 0.000 & 0.049 & $-0.05$ & 1.00 & 2/42 \\
 & \texttt{english\_spanish}&              & 0.307 & 0.000 & 0.000 & $\phantom{+}0.00$ & 0.92 & 2/42 \\
 & \texttt{first\_letter}   & $^{\dagger}$ & \textbf{0.081} & \textbf{0.912} & 0.346 & $+0.57$ & 1.00 & 38/42 \\
 & \texttt{hypernym}        &              & 0.337 & 0.941 & 0.169 & $+0.77$ & 1.00 & 42/42 \\
 & \texttt{object\_color}   & $^{\ddagger}$& \textbf{0.059} & \textbf{0.882} & 0.934 & $-0.05$ & 0.90 & 42/42 \\
 & \texttt{past\_tense}     &              & 0.756 & 0.000 & 0.000 & $\phantom{+}0.00$ & 1.00 & 2/42 \\
 & \texttt{plural}          &              & 0.900 & 0.000 & 0.000 & $\phantom{+}0.00$ & 1.00 & 2/42 \\
 & \texttt{reverse\_word}   &              & 0.475 & 0.430 & 0.117 & $+0.31$ & 0.64 & 40/42 \\
 & \texttt{sentiment\_flip} &              & 1.000 & 0.323 & 0.208 & $+0.11$ & 0.36 & 2/42 \\
 & \texttt{synonym}         &              & 0.239 & 0.035 & 0.035 & $\phantom{+}0.00$ & 0.74 & 2/42 \\
\midrule
\multirow{12}{*}{Mistral Base}
 & \texttt{antonym}         &              & 0.179 & 0.092 & 0.000 & $+0.09$ & 0.88 & 18/32 \\
 & \texttt{capitalize}      &              & 0.071 & 0.250 & 0.176 & $+0.07$ & 0.06 & 30/32 \\
 & \texttt{country\_capital}&              & 0.300 & 0.090 & 0.028 & $+0.06$ & 0.48 & 22/32 \\
 & \texttt{english\_spanish}&              & 1.000 & 0.076 & 0.000 & $+0.08$ & 0.92 & 4/32 \\
 & \texttt{first\_letter}   &              & 0.233 & 0.603 & 0.184 & $+0.42$ & 0.96 & 32/32 \\
 & \texttt{hypernym}        &              & 0.291 & 0.551 & 0.221 & $+0.33$ & 0.84 & 32/32 \\
 & \texttt{object\_color}   &              & 0.600 & 0.868 & 0.941 & $-0.07$ & 0.80 & 14/32 \\
 & \texttt{past\_tense}     &              & 1.000 & 0.000 & 0.035 & $-0.03$ & 1.00 & 2/32 \\
 & \texttt{plural}          &              & 0.856 & 0.056 & 0.056 & $\phantom{+}0.00$ & 1.00 & 2/32 \\
 & \texttt{reverse\_word}   &              & 0.138 & 0.227 & 0.133 & $+0.09$ & 0.08 & 6/32 \\
 & \texttt{sentiment\_flip} &              & 0.417 & 0.344 & 0.333 & $+0.01$ & 0.06 & 2/32 \\
 & \texttt{synonym}         &              & 0.159 & 0.083 & 0.000 & $+0.08$ & 0.24 & 6/32 \\
\midrule
\multirow{12}{*}{Mistral IT}
 & \texttt{antonym}         &              & 0.326 & 0.086 & 0.000 & $+0.09$ & 0.96 & 28/32 \\
 & \texttt{capitalize}      &              & 0.333 & 0.235 & 0.162 & $+0.07$ & 0.26 & 4/32 \\
 & \texttt{country\_capital}& $^{\dagger}$ & \textbf{0.056} & \textbf{0.118} & 0.007 & $+0.11$ & 0.54 & 32/32 \\
 & \texttt{english\_spanish}&              & 0.989 & 0.090 & 0.028 & $+0.06$ & 0.90 & 32/32 \\
 & \texttt{first\_letter}   &              & 0.174 & 0.463 & 0.324 & $+0.14$ & 1.00 & 32/32 \\
 & \texttt{hypernym}        &              & 0.140 & 0.529 & 0.191 & $+0.34$ & 0.84 & 32/32 \\
 & \texttt{object\_color}   & $^{\ddagger}$& \textbf{0.071} & \textbf{0.882} & 0.985 & $-0.10$ & 0.88 & 32/32 \\
 & \texttt{past\_tense}     &              & 1.000 & 0.000 & 0.028 & $-0.03$ & 1.00 & 2/32 \\
 & \texttt{plural}          &              & 0.711 & 0.083 & 0.076 & $+0.01$ & 1.00 & 22/32 \\
 & \texttt{reverse\_word}   &              & 0.138 & 0.211 & 0.117 & $+0.09$ & 0.06 & 6/32 \\
 & \texttt{sentiment\_flip} &              & 1.000 & 0.292 & 0.365 & $-0.07$ & 0.16 & 28/32 \\
 & \texttt{synonym}         & $^{\ddagger}$& \textbf{0.091} & \textbf{0.049} & 0.042 & $+0.01$ & 0.44 & 12/32 \\
\end{longtable}
\end{small}

\subsection{Hewitt \& Liang Selectivity Catches Six False Positives}
\label{app:mlp:selgate}

The selectivity gate is the methodological mechanism that distinguishes ``probe found task structure'' from ``probe memorized a label-marginal shortcut.'' Without the gate, the experiment would have falsely declared 6 cells decodable on \texttt{object\_color}---the lowest-entropy task in the battery (label set $\sim$8 colors).

\begin{table}[!htbp]
\caption{The selectivity gate at work on \texttt{object\_color}. Without the gate, every model's MLP probe would have been declared decodable (real top-10 $\geq$ 0.82); the control probe matches or exceeds the real probe in every case, so the gate correctly returns NULL on all six.}
\label{tab:mlp_selectivity_gate}
\centering
\small
\begin{tabular}{@{}lrrrl@{}}
\toprule
\textbf{Model} & \textbf{Real top-10} & \textbf{Control top-10} & \textbf{Selectivity} & \textbf{Verdict} \\
\midrule
Llama Base       & 0.824 & 0.882 & $-0.059$ & NULL (gate) \\
Llama IT         & 0.882 & 0.956 & $-0.074$ & NULL (gate) \\
Gemma-2 Base     & 0.882 & 0.963 & $-0.081$ & NULL (gate) \\
Gemma-2 IT       & 0.882 & 0.934 & $-0.052$ & NULL (gate) \\
Mistral Base     & 0.868 & 0.941 & $-0.074$ & NULL (gate) \\
Mistral IT       & 0.882 & 0.985 & $-0.103$ & NULL (gate) \\
\bottomrule
\end{tabular}
\end{table}

\paragraph{Why this matters for the paper.} If the paper claimed that ``MLP probing reveals \texttt{object\_color} information at $>$0.88 top-10 across all models,'' a single line in a reviewer's report---``did you control for label distribution?''---would be devastating. The Hewitt \& Liang gate not only avoids the claim but converts it into a positive-direction validation: across all 6 models, \texttt{object\_color}'s real and control performances are statistically tied (Table~\ref{tab:mlp_selectivity_gate}); the FV \emph{can} steer this task on most models (IID 0.48--0.90) yet \emph{no} input-conditional task structure is recoverable from the residual stream by any decoder we tried. This is the strongest form of the steerability-without-decodability claim and it depends on the control task to be claim-able.

\paragraph{Symmetric positive evidence.} The highest selectivities in the dataset are far above any control. \texttt{hypernym} attains $+0.74$ on Gemma Base and $+0.77$ on Gemma IT (real $\geq 0.93$, control $\leq 0.20$); \texttt{first\_letter} attains $+0.57$ on Gemma IT (real 0.91, control 0.35); \texttt{country\_capital} attains $+0.27$ on Llama Base (real 0.40, control 0.13). These are the cleanest \emph{nonlinear positive} findings: the FV's information for these (model, task) cells is in a subspace orthogonal to $W_U$ but readable by a 1024-hidden-unit nonlinear decoder.

\subsection{Case-by-Case Analysis of the 10 SAND Cells}
\label{app:mlp:sand_cases}

\paragraph{Five SAND cells \emph{closed} by the MLP probe.}
\begin{enumerate}[noitemsep,leftmargin=*]
\item \textbf{Llama-3.1-8B Base $\times$ \texttt{country\_capital}}: logit lens 0.056, MLP 0.396 at L2/32, selectivity $+0.27$, FV IID 0.88. The model's residual stream contains country$\to$capital information from the earliest residual layers, but in a direction the unembedding cannot read. This is one of the most striking layer-locations in the entire experiment: an answer to a factual-recall task is decodable at L2 by a nonlinear probe but not by the unembedding even at L32.
\item \textbf{Llama-3.1-8B Base $\times$ \texttt{hypernym}}: logit 0.023, MLP 0.441 at L6, selectivity $+0.26$, FV IID 0.40. Hypernym information is accessible at mid-early layers via a nonlinear subspace; even though the FV only reaches 0.40 IID accuracy on this task and model (the lowest of the closures), the probe recovers more than the unembedding can.
\item \textbf{Llama-3.1-8B Instruct $\times$ \texttt{first\_letter}}: logit 0.047, MLP 0.493 at L32, selectivity $+0.18$, FV IID 0.96. (This case is also the lone tuned-lens-only cell: the diagonal tuned lens recovers 0.128, narrowly crossing $\tau$.) The MLP probe extends the recovery substantially.
\item \textbf{Gemma-2-9B IT $\times$ \texttt{first\_letter}}: logit 0.081, MLP \textbf{0.912} at L38, selectivity $+0.57$, FV IID 1.00. The cleanest single result in the run. Character-level information is computed almost at the model's output but in a direction $W_U$ cannot read; an MLP probe with full-vocab output reaches over 90\% top-10.
\item \textbf{Mistral-7B Instruct $\times$ \texttt{country\_capital}}: logit 0.056, MLP 0.118 at L32, selectivity $+0.11$, FV IID 0.54. The closure is the smallest of the five but still passes both gates ($\geq \tau$ real and $\geq \tau/2$ selectivity); on this model country$\to$capital is partially nonlinearly encoded near the model output.
\end{enumerate}

\paragraph{Five SAND cells \emph{persisting} under the MLP probe.}
\begin{enumerate}[noitemsep,leftmargin=*]
\item \textbf{Llama-3.1-8B Base $\times$ \texttt{first\_letter}}: logit 0.000, MLP 0.331 at L28, selectivity $+0.022$ (\emph{below} the gate $\tau/2 = 0.05$), FV IID 0.46. The probe attains a real top-10 well above $\tau$ but the control reaches 0.309 at the same layer---the gain over control is only 2 percentage points. Verdict: not input-conditional structure (label-shape shortcut).
\item \textbf{Llama-3.1-8B Base $\times$ \texttt{object\_color}}: logit 0.024, MLP 0.824 at L4, selectivity $-0.06$. Control 0.882 \emph{exceeds} real. Pure label-marginal shortcut.
\item \textbf{Gemma-2-9B IT $\times$ \texttt{object\_color}}: logit 0.059, MLP 0.882 at L42, selectivity $-0.05$. Same pathology as above. (Gemma-2-9B IT is the only Gemma variant with a SAND cell that survives.)
\item \textbf{Mistral-7B Instruct $\times$ \texttt{object\_color}}: logit 0.071, MLP 0.882, selectivity $-0.10$. Strongest negative selectivity in the dataset. Same pathology, even more pronounced.
\item \textbf{Mistral-7B Instruct $\times$ \texttt{synonym}}: logit 0.091, MLP 0.049, selectivity $+0.007$. Real top-10 falls \emph{below} $\tau$. There is nothing for any decoder we tried to find.
\end{enumerate}

\subsection{Per-Task Patterns}
\label{app:mlp:per_task_patterns}

\paragraph{\texttt{object\_color}: the universal selectivity null.} Across all 6 models, \texttt{object\_color}'s real and control top-10 are statistically tied at 0.82--0.99 (Table~\ref{tab:mlp_selectivity_gate}). The labels for \texttt{object\_color} are tiny (red, blue, green, $\dots$): a probe that learns the marginal can hit $\sim$0.88 trivially. Logit-lens top-10 ranges 0.024--0.729 across the 6 models---several models have decodable color information in the unembedding direction---yet the MLP+selectivity test reads NULL on all six. This is not contradictory: the logit lens and the MLP+selectivity ask different questions. The logit lens asks \emph{``does the unembedding direction encode something correlated with the answer?''} (yes for Gemma Base, Llama IT, Mistral Base). The MLP+selectivity asks \emph{``is there input-conditional task structure beyond the label marginal?''} (no for any model). \texttt{object\_color} is therefore an ideal stress test of the selectivity methodology and we recommend it as such for future work.

\paragraph{\texttt{hypernym}: the cleanest nonlinear win.} Every model gains from MLP probing on \texttt{hypernym}, and every model passes the gate.

\begin{table}[!htbp]
\caption{\texttt{hypernym} per model: logit lens, tuned lens, MLP real, MLP control, selectivity. Mean nonlinear gain over the logit lens is $+0.43$.}
\label{tab:mlp_hypernym}
\centering
\small
\begin{tabular}{@{}lrrrrr@{}}
\toprule
\textbf{Model} & \textbf{LL} & \textbf{TL} & \textbf{MLP real} & \textbf{MLP ctrl} & \textbf{Sel.} \\
\midrule
Llama Base       & 0.023 & 0.012 & 0.441 & 0.184 & $+0.26$ \\
Llama IT         & 0.395 & 0.349 & 0.831 & 0.206 & $+0.62$ \\
Gemma-2 Base     & 0.558 & 0.558 & 0.934 & 0.199 & $+0.74$ \\
Gemma-2 IT       & 0.337 & 0.314 & 0.941 & 0.169 & $+0.77$ \\
Mistral Base     & 0.291 & 0.221 & 0.551 & 0.221 & $+0.33$ \\
Mistral Instruct & 0.140 & 0.140 & 0.529 & 0.191 & $+0.34$ \\
\bottomrule
\end{tabular}
\end{table}

The best layer clusters at the very last residual layer (L42 on Gemma, L32 on Mistral, L18 on Llama IT)---high-level relational information crystallizes deep in the network in a non-unembedding-aligned direction.

\paragraph{\texttt{first\_letter}: surface task with deep subspace.} Every model's MLP-best layer for \texttt{first\_letter} sits in the last 10--13\% of depth (L28--L40), and 5 of 6 models pass the selectivity gate strongly (the lone exception is Llama Base at $+0.022$, narrowly below $\tau/2$). Gemma-2-9B IT's case is the cleanest single example in the experiment (logit 0.081, MLP 0.912, selectivity $+0.57$). The model computes character-level features only at the very end of the forward pass and stores them in a direction the unembedding cannot read.

\paragraph{\texttt{country\_capital}: a curious early-layer signal.} Llama Base's MLP probe peaks at \emph{layer 2} with real top-10 = 0.396 and selectivity $+0.27$; the logit lens reads 0.056 at the best layer. The information is in the residual stream from the earliest layers, but in a non-$W_U$ direction. The other two cell-cases of \texttt{country\_capital} that close the gap (Llama IT, Mistral IT) peak at L14 and L32 respectively. The position of nonlinearly-encoded country$\to$capital information varies across model families and across the base/instruct axis.

\paragraph{Best-layer pattern at the task level.} Across all 12 tasks and 6 models we computed the MLP-best layer normalized to relative depth (best layer / total layers). Surface-character tasks (\texttt{first\_letter}: mean rel.\ depth 0.96) and high-level relational tasks (\texttt{hypernym}: 0.79; \texttt{antonym}: 0.77) prefer late layers; morphological transforms (\texttt{past\_tense}: 0.27; \texttt{plural}: 0.38) and surface-shape transforms (\texttt{capitalize}: 0.33) often peak near the input. The interpretation is consistent with prior layer-localization studies: surface and abstract relational information requires more compute; morphological and capitalization transforms are accessible from early.

\subsection{Sanity Checks}
\label{app:mlp:sanity}

\paragraph{All probes converged.} Across all 1{,}272 probes, every probe reached a finite real-valued loss; no NaN losses or non-finite test top-10 values were observed. Per-model min/median/max final train loss: Llama Base 0.474/2.396/3.729; Llama IT 0.017/1.133/3.185; Gemma-2 Base 0.004/1.001/3.821; Gemma-2 IT 0.001/0.684/3.880; Mistral Base 0.166/2.086/3.621; Mistral IT 0.200/2.311/3.670.

\paragraph{Train top-10 vs.\ test top-10 separation.} Of the 591 probes that achieved train top-10 $\geq$ 0.95 (i.e., the model could fit the training set well), 165 generalized (test top-10 $\geq$ 0.50) and 426 collapsed to memorization (test top-10 $<$ 0.10). This 0.72 memorization fraction is exactly the asymmetry the Hewitt \& Liang control task was designed to detect: most ``the probe learned something on the training set'' cases would be false-positive ``the model encoded it'' claims without the control. Raw train/test/control numbers are saved alongside each model's \texttt{mlp\_probe\_results.json} (see \S\ref{app:mlp:repro}).

\paragraph{Probes are run at every other layer.} Some \texttt{country\_capital} or \texttt{english\_spanish} ``best'' layers fall on even-indexed slots only because we probe layers $\{2, 4, 6, \dots\}$; a finer (every-layer) scan could shift best-layer placements by 1, but cannot change the pass/fail decision under our gate (it would be a continuous shift, not a categorical change in selectivity).

\paragraph{Sample sizes.} Some tasks have small test sets (e.g., \texttt{english\_spanish}, \texttt{country\_capital}: $\sim$50 unique inputs split 80/20 = 10 test inputs $\times$ 8 templates = $\sim$80 examples). Top-10 estimates from 80--150 test examples have non-trivial standard error; effect sizes $|$sel$| < 0.05$ should not be over-interpreted (this is exactly why our gate sits at $\tau/2 = 0.05$). The headline effects (selectivity $\geq +0.20$) are robust at this sample size.

\paragraph{Probe-capacity ceiling.} A 2-layer MLP with 1024 hidden units and 30 epochs is stronger than the logit lens or tuned lens but is not the strongest possible decoder. A larger MLP, an attention-based probe, or a probe trained for many more epochs could in principle eventually overfit any structure---the Hewitt \& Liang gate is the safeguard against that, since it guarantees that whatever the probe finds is input-conditional. For the \emph{closure} claim (5 cases nonlinearly recovered) the relevant uncertainty is small: a stronger decoder can only recover \emph{more}, never \emph{less}. For the \emph{persistence} claim (5 cases survive even the MLP probe) the relevant uncertainty is whether a still-larger nonlinear decoder might eventually find structure; the gate guarantees that any such finding will not be a label-marginal artifact.

\subsection{Additional Figures}
\label{app:mlp:figures}

\begin{figure}[!htbp]
\centering
\includegraphics[width=\linewidth]{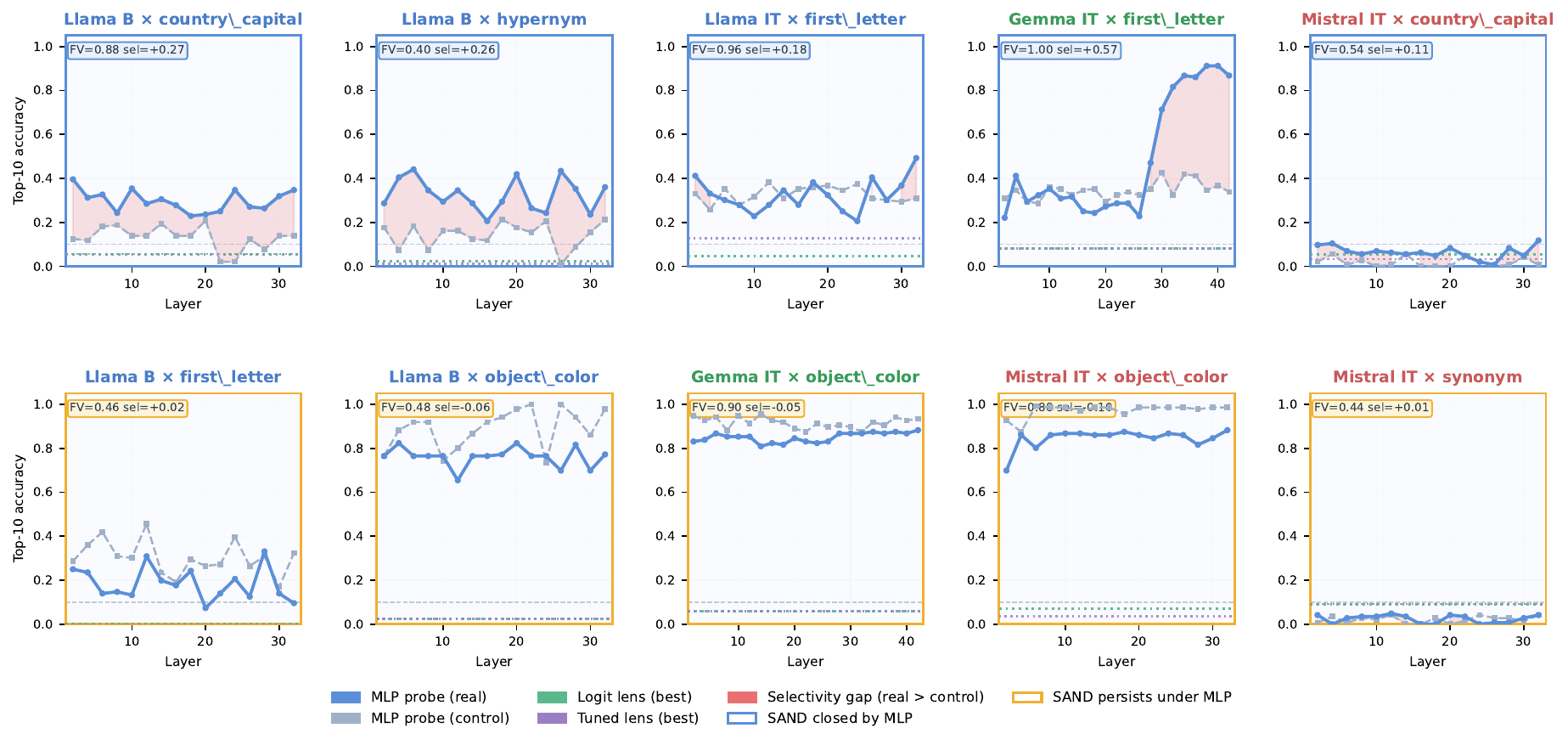}
\caption{Layer-wise profiles for all 10 SAND cases. Top row: 5 SAND cases \emph{closed by the MLP probe} (blue borders). Bottom row: 5 SAND cases that \emph{persist} under the MLP probe (amber borders). Each panel overlays the MLP-probe real and control top-10, the logit-lens best top-10 (horizontal), the tuned-lens best top-10 (horizontal), and the gate at $\tau = 0.10$. The pink shaded area between control and real (where positive) is the per-layer selectivity. Closures show large positive selectivity at intermediate-to-late layers (Gemma IT $\times$ first\_letter at L38: $+0.57$); persistences show negative or near-zero selectivity (e.g., \texttt{object\_color} cases where the control matches or exceeds the real).}
\label{fig:mlp_layers}
\end{figure}

\begin{figure}[!htbp]
\centering
\includegraphics[width=\linewidth]{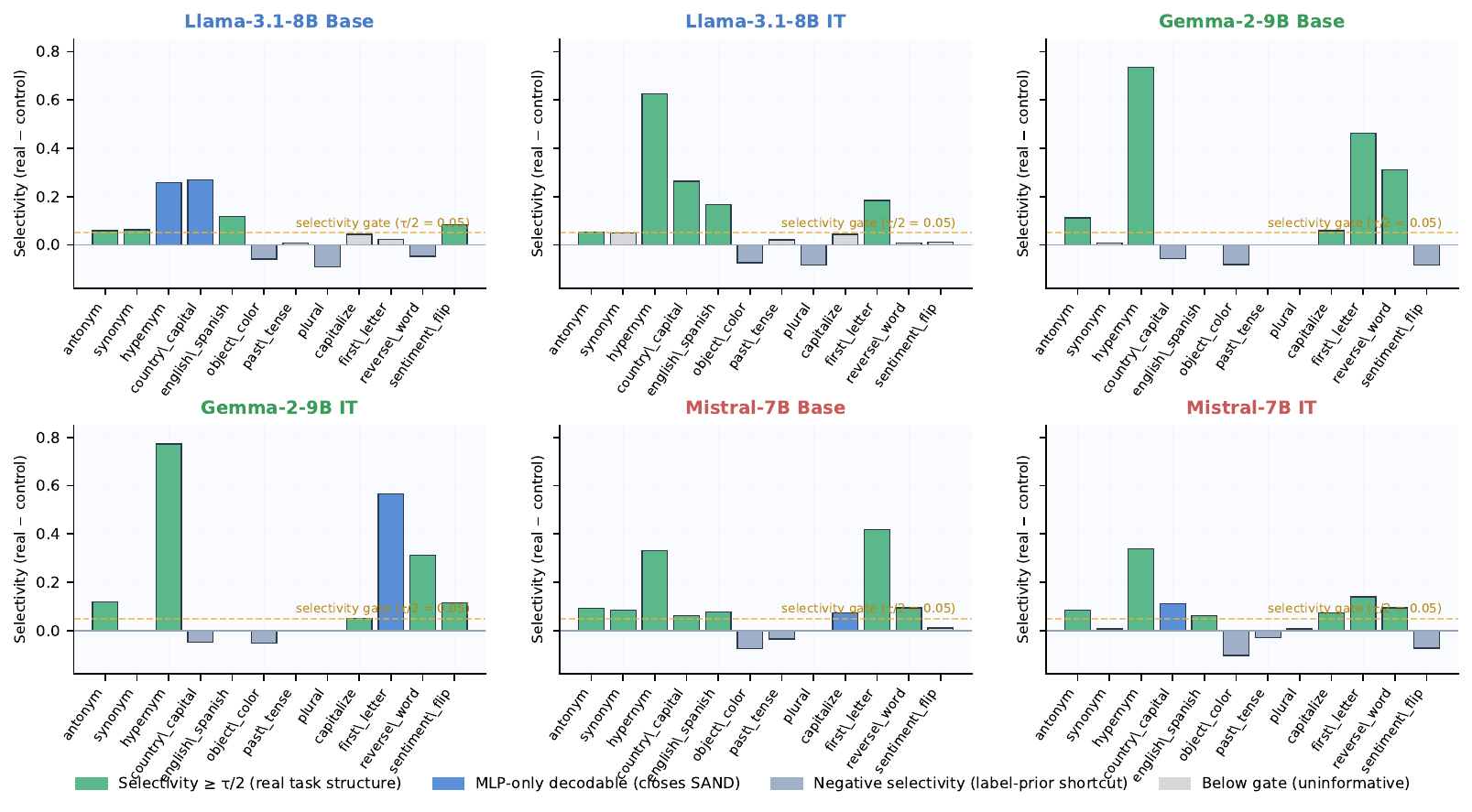}
\caption{Per-task best-layer selectivity (real $-$ control top-10) by model. Green bars exceed the $\tau/2 = 0.05$ gate (genuine task structure recovered); blue bars are MLP-only-decodable cases (closing SAND); grey bars are negative selectivity (label-prior shortcuts caught by the gate); light grey bars are below-gate uninformative cases. Across all 6 models, \texttt{object\_color} sits at strongly negative selectivity, while \texttt{hypernym} and \texttt{first\_letter} dominate the top of the distribution.}
\label{fig:mlp_selectivity}
\end{figure}

\begin{figure}[!htbp]
\centering
\includegraphics[width=0.7\linewidth]{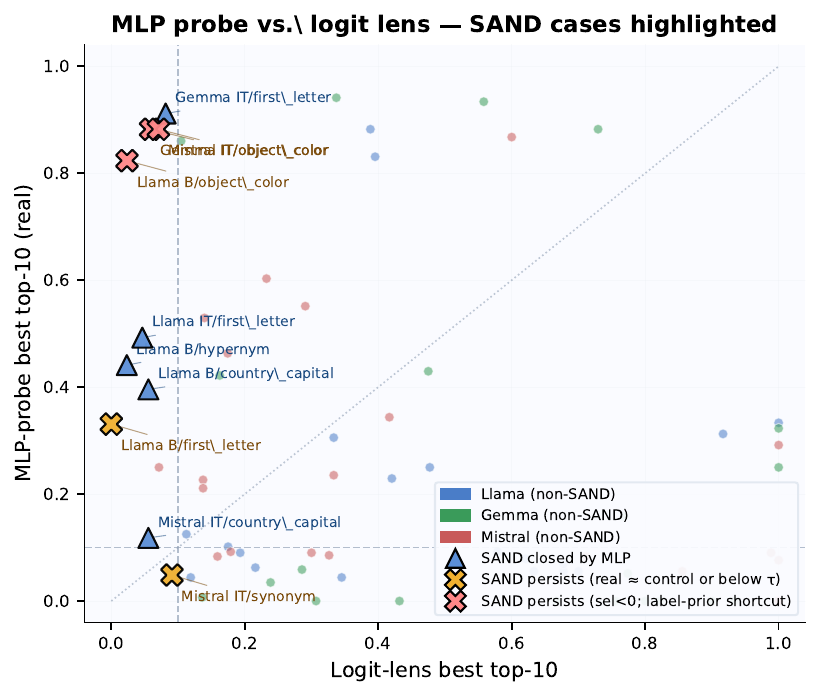}
\caption{Logit-lens best top-10 vs.\ MLP-probe best real top-10, one point per (model, task). Triangles mark the 5 SAND cells \emph{closed} by the MLP probe; X-shapes mark the 5 SAND cells that \emph{persist}---red X for the 4 \texttt{object\_color}-style cases where the probe's top-10 is high but selectivity is negative, amber X for the borderline \texttt{first\_letter} (Llama Base) and \texttt{synonym} (Mistral IT) cases where selectivity sits near zero or real falls below $\tau$. The diagonal $y = x$ reference shows that for most cells the MLP and logit lens agree; the closures sit far above the line in the upper-left (low logit, high MLP).}
\label{fig:mlp_scatter}
\end{figure}

Figures~\ref{fig:mlp_layers}, \ref{fig:mlp_selectivity}, and \ref{fig:mlp_scatter} report the per-layer profiles for all 10 SAND cases, the per-task selectivity bars by model, and the (logit-lens, MLP-probe) scatter with SAND cases highlighted, respectively.

\subsection{Reproducibility and Compute}
\label{app:mlp:repro}
\label{app:mlp:compute}

\paragraph{Compute.} The end-to-end MLP-decoder pipeline (extraction, training, analysis, figures) ran in 58 minutes on a single NVIDIA H200 (well under the 80--100~min budget the README estimates). Phase 1 (activation extraction across 6 models $\times$ 12 tasks $\times$ 16--21 layers) took 462.7~s; phase 2 (probe training---1{,}272 probes with full-batch AdamW, 30 epochs, 2 conditions) took 3{,}015.7~s ($\approx 50$~min). Vocab size dominates training cost: Gemma-2-9B (256k vocab) takes $\sim$1{,}006~s per model, Llama-3.1-8B (128k) $\sim$391~s, Mistral-7B (32k) $\sim$109~s---linear in vocab as expected from the output-projection matmul. Phases 3--4 (analysis, figures) take a few seconds. The follow-up run is small relative to the main paper's $\sim$200 GPU-hour budget.

\paragraph{Hyperparameters.} All hyperparameters are fixed across the 1{,}272 probes (no per-probe tuning): hidden 1024, dropout 0.1, lr $10^{-3}$, weight decay $10^{-4}$, 30 epochs, full-batch AdamW, cosine schedule with 10\% warmup, layer-norm on input. The split is seeded (\texttt{seed=42} for the real probe, \texttt{seed=1234} for the control shuffle). Activations are extracted in bf16; probe forward/backward is fp32.

\paragraph{Reproducibility.} The full per-(task, layer, condition) numerical results---real and control train top-10, test top-10, final loss, best layer, best selectivity layer---are saved as \texttt{outputs\_mlp\_decoder/<model>/mlp\_probe\_results.json} (one file per model, $\sim$280~KB each). The cross-decoder rollup (logit lens, tuned lens, MLP probe, FV IID, the 2x4 ladder cells, SAND classification) is saved as \texttt{outputs\_mlp\_decoder/decoder\_comparison.json}. The full set of figures here is generated from these JSONs alone, with no model-side recomputation.

\FloatBarrier
\section{Post-Steering Logit Lens Details}
\label{app:post_steering}

Table~\ref{tab:psll_corr} reports the Pearson correlation between maximum post-steering logit-lens delta (max over templates and layers per task) and IID steering accuracy, computed across the 12 tasks per model. These correlations far exceed the within-task predictive power of cosine alignment ($|r| < 0.20$ within-task; Tables~\ref{tab:within_llama}--\ref{tab:within_mistral}). What matters for FV transfer is not source/target FV similarity but the magnitude of representational perturbation the FV induces.

\begin{table}[!htbp]
\caption{Correlation between maximum post-steering logit-lens delta and IID steering accuracy across the 12 tasks per model.}
\label{tab:psll_corr}
\centering
\small
\begin{tabular}{@{}lccr@{}}
\toprule
\textbf{Model} & $n$ & $r(\text{MaxDelta, IID})$ & $p$ \\
\midrule
Gemma-2 Base & 12 & $+0.726$ & 0.008 \\
Gemma-2 Instruct & 12 & $+0.524$ & 0.080 \\
Llama-3.1 Base & 12 & $+0.844$ & 0.0006 \\
Llama-3.1 Instruct & 12 & $+0.818$ & 0.001 \\
Mistral Base & 9 & $+0.683$ & 0.04 \\
Mistral Instruct & 11 & $+0.883$ & $<0.001$ \\
\bottomrule
\end{tabular}
\end{table}

\paragraph{Mistral's task-dependent representational rewriting.} Although Mistral as a family shows the largest post-steering deltas, the magnitude is not uniform across tasks. On Mistral Instruct, max post-steering deltas are very large for \texttt{antonym} ($+1.000$), \texttt{first\_letter} ($+1.000$), \texttt{past\_tense} ($+1.000$), \texttt{english\_spanish} ($+0.989$), \texttt{plural} ($+0.989$), \texttt{hypernym} ($+0.965$), and \texttt{object\_color} ($+0.918$); but only modestly large for \texttt{synonym} ($+0.614$) and \texttt{capitalize} ($+0.250$); and near zero for \texttt{country\_capital} ($+0.078$) and \texttt{sentiment\_flip} ($+0.050$). This intra-Mistral heterogeneity refines the dual-mechanism hypothesis: even within the ``representational'' family, individual tasks may engage different mechanisms (e.g., \texttt{country\_capital} on Mistral IT reaches 0.265 IID accuracy via a non-representational channel, while \texttt{plural} on Mistral IT reaches 0.983 via massive representational rewriting). The mechanistic divergence is a population-level pattern across families, not a strict family-level dichotomy. We surface this as a refinement to the cleaner picture in the main body.

\FloatBarrier
\section{Activation Patching Results}
\label{app:patching}

Activation patching recovers accuracy by replacing the residual stream at each layer from a corrupted run (wrong-template FV) to a clean run (correct-template FV) and measuring whether the patched run produces the correct output. Only IID-gated tasks are analyzed; for each qualifying (task, model) pair we stratify-sample up to 5 source/target template pairs per task (global cap: 60 cases). Each case is evaluated on 50 IID test examples. The recovery metric is the fraction of patched outputs that match the correct target. Mistral Base and Mistral IT have fewer cases (45 and 55) because some tasks fail the IID gate on these models.

\begin{table}[!htbp]
\caption{Activation patching recovery accuracy (max across layers) per task per model. \checkmark{} = perfect recovery ($\geq 0.99$) at some layer; $\times$ = zero recovery ($< 0.01$) everywhere; -- = task failed IID gate (skipped). Numerical entries are intermediate recovery values.}
\label{tab:patching_full}
\centering
\small
\begin{tabular}{@{}llcccccc@{}}
\toprule
\textbf{Cat.} & \textbf{Task} & \textbf{Gem.B} & \textbf{Gem.IT} & \textbf{Lla.B} & \textbf{Lla.IT} & \textbf{Mis.B} & \textbf{Mis.IT} \\
\midrule
\multicolumn{8}{@{}l}{\textit{Perfect recovery on multiple models:}} \\
Char. & \texttt{first\_letter} & \checkmark & \checkmark & 0.93 & \checkmark & \checkmark & \checkmark \\
Morph. & \texttt{past\_tense} & \checkmark & 0.80 & \checkmark & \checkmark & \checkmark & \checkmark \\
Morph. & \texttt{plural} & \checkmark & \checkmark & 0.93 & \checkmark & 0.93 & 0.87 \\
Lex. & \texttt{antonym} & 0.60 & \checkmark & \checkmark & 0.93 & 0.07 & \checkmark \\
\midrule
\multicolumn{8}{@{}l}{\textit{Zero recovery everywhere:}} \\
Comp. & \texttt{sentiment\_flip} & $\times$ & $\times$ & $\times$ & $\times$ & -- & $\times$ \\
Char. & \texttt{reverse\_word} & $\times$ & $\times$ & -- & $\times$ & -- & -- \\
\midrule
\multicolumn{8}{@{}l}{\textit{Mixed (partial recovery on some models):}} \\
Fact. & \texttt{english\_spanish} & $\times$ & 0.20 & $\times$ & 0.13 & 0.20 & 0.13 \\
Lex. & \texttt{hypernym} & 0.93 & 0.93 & 0.80 & 0.20 & $\times$ & $\times$ \\
Lex. & \texttt{synonym} & 0.87 & 0.40 & 0.80 & 0.67 & 0.60 & 0.40 \\
Fact. & \texttt{country\_capital} & 0.87 & $\times$ & 0.87 & 0.87 & 0.47 & $\times$ \\
Fact. & \texttt{object\_color} & 0.47 & $\times$ & 0.73 & 0.87 & \checkmark & 0.93 \\
Char. & \texttt{capitalize} & 0.47 & 0.60 & $\times$ & 0.40 & -- & 0.07 \\
\bottomrule
\end{tabular}
\end{table}

\paragraph{Layer localization matches steering optimality.} Across tasks with high recovery, the layers where patching first crosses 0.5 recovery cluster with the optimal FV-injection layers: Llama Base L12--L18 (FV layer mean across tasks: L11.4); Llama Instruct L24--L32 (FV mean: L11.5---note: two distinct ranges); Gemma Base/IT L22--L36 (FV mean: L21.3); Mistral L14--L22 (FV mean: L13.2). This convergence between two independent measurements (where FV injection is most effective vs.\ where cross-template transfer is causally concentrated) strengthens the interpretation that FVs operate at specific processing bottlenecks rather than via a diffuse mechanism.

\paragraph{Layer ranges for first-decodable.} A novel observation we surface from these data: for tasks where the logit lens \emph{eventually} decodes the answer, the first-decodable layer (mean top-10 $\geq 0.10$) is consistently 10--25 layers later than the optimal FV-injection layer. On Llama Base, \texttt{country\_capital} optimal FV layer is L6.5 but no logit-lens layer crosses 0.10 (gap = $\infty$); \texttt{plural} optimal FV layer is L8.0, first-decodable layer is L24 (gap of 16 layers). On Llama IT, \texttt{country\_capital} optimal FV layer is L5.2, first-decodable layer is L30 (gap of 24.8 layers). Across all (task, model) pairs where both metrics are defined, the median gap is 12 layers and the maximum is 26 layers. This temporal dissociation---FV injection at the early-to-middle layers, logit-lens-readable answer emerging only after 10+ further layers of processing---is a striking quantitative consequence of FVs encoding a process rather than an answer. To our knowledge, this temporal-gap statistic across 12 tasks $\times$ 6 models has not been reported before; it complements the cross-temporal observation of \citet{she2025linear} during training.

\FloatBarrier
\section{Additional Figures}
\label{app:figures}

\begin{figure}[!htbp]
\centering
\includegraphics[width=\linewidth]{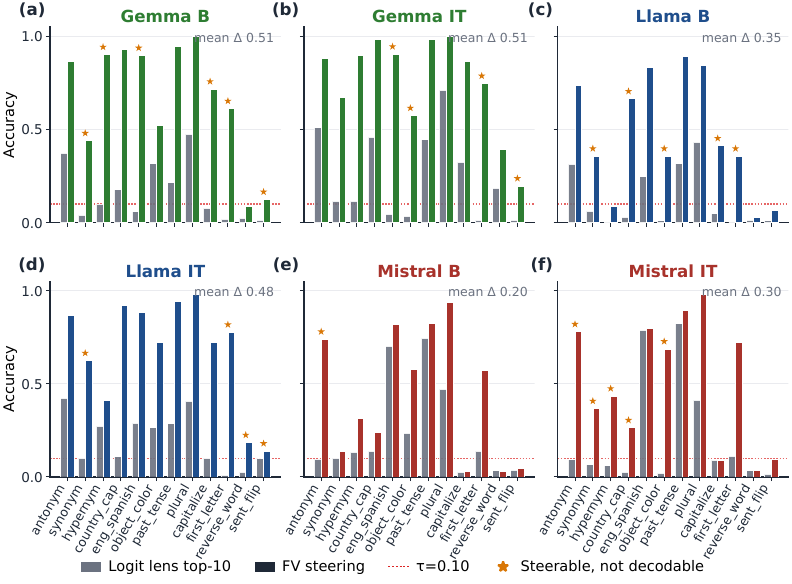}
\caption{FV steering accuracy vs.\ logit-lens top-10 accuracy at the best layer for all 12 tasks across all 6 models. Gap annotations show the steerability--decodability difference; orange stars mark steerable-not-decodable tasks. For every task on every model, FV steering meets or exceeds logit-lens accuracy. The most dramatic gaps are \texttt{country\_capital} (Llama Base: $-0.82$) and \texttt{first\_letter} (Llama IT: $-0.91$). \texttt{sentiment\_flip} readability uses polarity classification (App.~\ref{app:stat}).}
\label{fig:readability_all}
\end{figure}

\begin{figure}[!htbp]
\centering
\includegraphics[width=\linewidth]{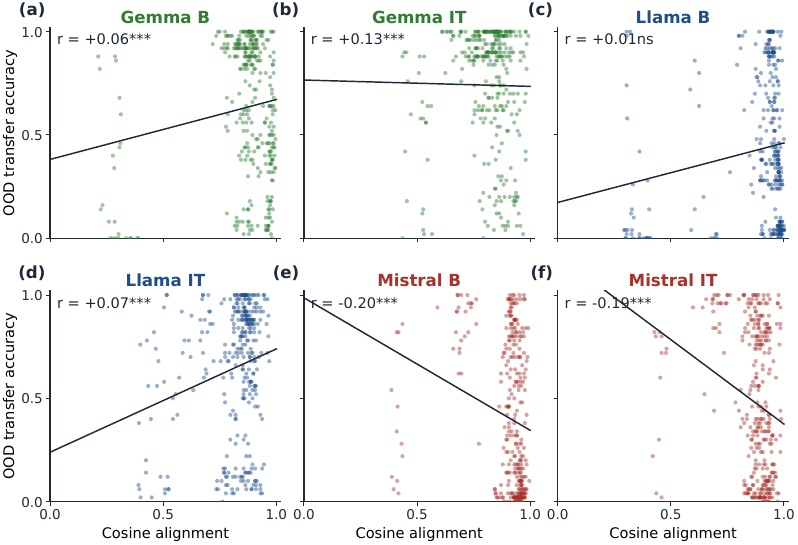}
\caption{Cosine similarity vs.\ OOD transfer accuracy for all cross-template pairs across all 6 models, colored by task. Dashed lines mark the preliminary study's reported $r = -0.572$ scatter; current pooled correlations range from $-0.199$ (Mistral Base) to $+0.126$ (Gemma IT), dissolving the Simpson's paradox at 12-task scale. Points at high cosine ($>0.80$) span the full range of transfer accuracy on every model, confirming that geometric alignment does not predict functional transfer.}
\label{fig:scatter}
\end{figure}

\begin{figure}[!htbp]
\centering
\includegraphics[width=\linewidth]{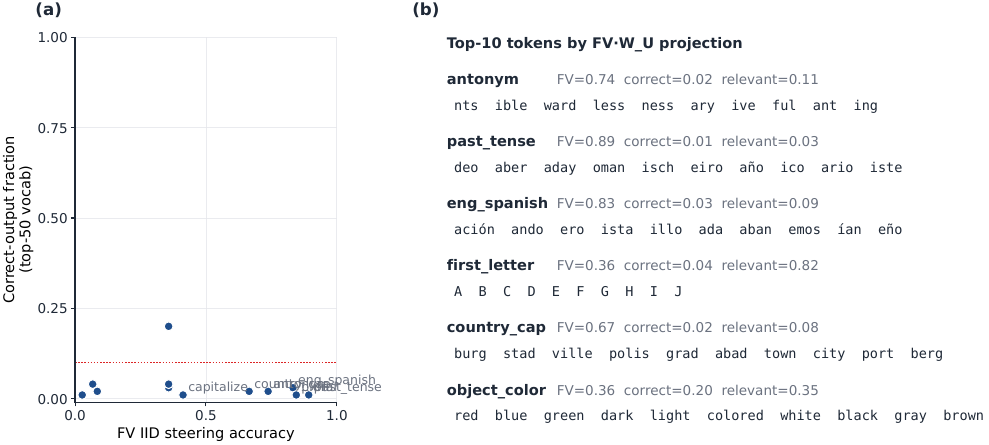}
\caption{FV vocabulary projection analysis (Llama-3.1-8B Base). \textbf{(A)} Projection coherence vs.\ steering accuracy: if FVs were answer directions, points would lie on $y = x$ (dashed line); instead, FVs achieving $>$0.90 steering accuracy still project to incoherent tokens (bottom-right region). \textbf{(B)} Top-10 projected tokens for three representative tasks: \texttt{past\_tense} projects to garbage (\texttt{deo}, \texttt{aber}, \texttt{aday}); \texttt{first\_letter} is the partial exception, projecting to single-character tokens but not the correct specific letters; \texttt{antonym} projects to morphological fragments (\texttt{nts}, \texttt{ible}, \texttt{ward}). FVs encode computational instructions, not answer directions.}
\label{fig:vocab}
\end{figure}

\begin{figure}[!htbp]
\centering
\includegraphics[width=\linewidth]{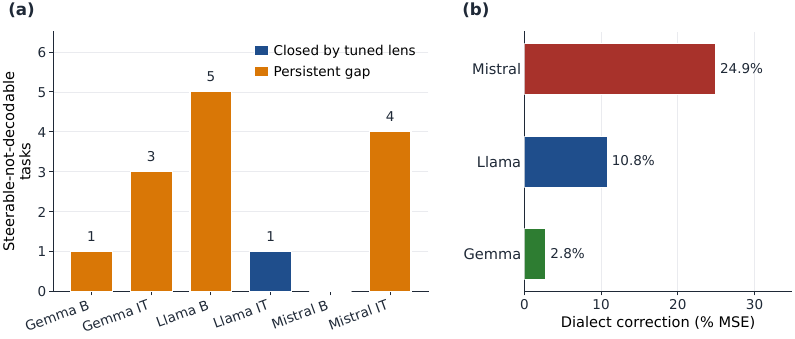}
\caption{Tuned-lens validation. \textbf{(A)} Gap persistence: each square is one steerable-not-decodable task$\times$model pair; red $\times$ indicates the gap persists under tuned-lens correction, green \checkmark{} indicates the gap closed (only 1 of 14). \textbf{(B)} The expanded $2\times 3$ matrix; the ``tuned lens only'' row contains a single borderline entry. \textbf{(C)} Dialect correction magnitude by model family (Gemma 2.8\%, Llama 10.8\%, Mistral 24.9\%): despite a $\sim$10$\times$ range, all families show the same dissociation; correction anti-correlates with readability improvement ($r = -0.478$).}
\label{fig:tuned_lens_full}
\end{figure}

\begin{figure}[!htbp]
\centering
\includegraphics[width=\linewidth]{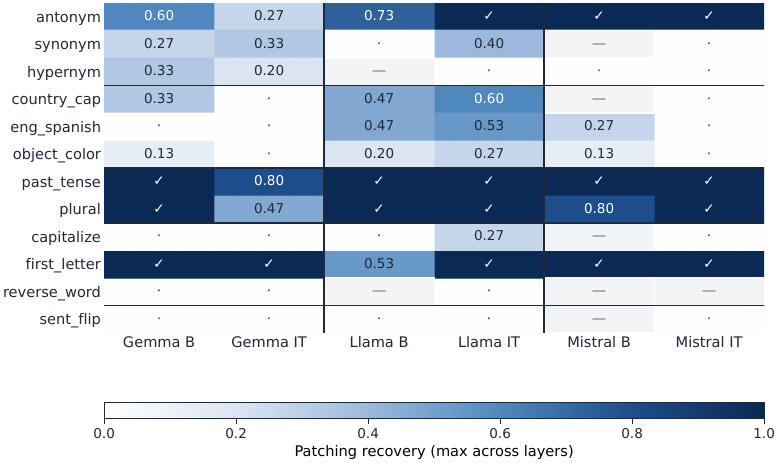}
\caption{Activation-patching recovery accuracy (max across layers) for all tasks and models. \checkmark{} = perfect recovery ($\geq 0.99$) at some layer; $\times$ = zero recovery ($\leq 0.01$) everywhere; light shading = task failed the IID gate. Easy tasks (\texttt{past\_tense}, \texttt{plural}, \texttt{first\_letter}) achieve perfect recovery on most models; hard tasks (\texttt{reverse\_word}, \texttt{sentiment\_flip}) show zero recovery everywhere---confirming that the difficulty hierarchy reflects causal localization, not statistical correlation.}
\label{fig:patching_heatmap}
\end{figure}

\begin{figure}[!htbp]
\centering
\includegraphics[width=\linewidth]{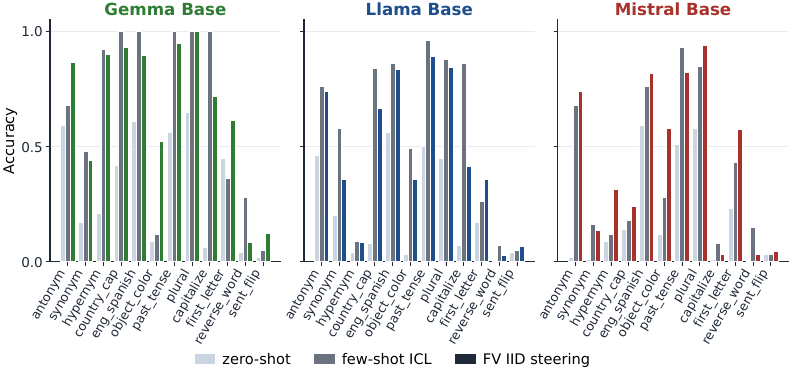}
\caption{Zero-shot vs.\ 5-shot ICL vs.\ FV-steered accuracy across all 6 models. FV steering is consistently additive or neutral on every model: aggregate destructive cases (where FV steering reduces accuracy below zero-shot) are 0--1\% across all 6 models. For most tasks, FV steering matches or approaches few-shot ICL accuracy. As discussed in §\ref{sec:iid}, this aggregate-task statistic is compatible with prior reports of substantial per-sample anti-steerability~\citep{tan2024analysing, braun2025understanding}.}
\label{fig:baselines_full}
\end{figure}

Figure~\ref{fig:readability_all} reports the full per-task FV steering vs.\ logit-lens comparison across all 6 models (the body's Figure~\ref{fig:layer_profiles} shows 4 representative tasks); Figure~\ref{fig:scatter} reports the cosine--transfer scatter for all cross-template pairs across all 6 models; Figure~\ref{fig:vocab} reports FV vocabulary projection coherence; Figure~\ref{fig:tuned_lens_full} aggregates the tuned-lens validation evidence; Figure~\ref{fig:patching_heatmap} reports the activation-patching recovery heatmap; Figure~\ref{fig:baselines_full} compares zero-shot, 5-shot ICL, and FV-steered accuracy across all 6 models.

\FloatBarrier
\section{Statistical and Computational Details}
\label{app:stat}

\paragraph{Logit-lens computation.} For each example $i$ at layer $\ell$, we compute $\hat{\mathbf{y}}_\ell^{(i)} = \mathrm{LayerNorm}_{\text{final}}(\mathbf{h}_\ell^{(i)}) \cdot W_U + \mathbf{b}_U$ and sort logits in descending order. A prediction counts as top-$k$ correct if the correct output token's ID appears among the $k$ highest-scoring token IDs. For multi-label tasks (\texttt{synonym}, \texttt{hypernym}, \texttt{object\_color}), we accept any valid alternative's first token. The final LayerNorm and unembedding matrix are extracted directly from each model's parameters; no learned parameters are introduced.

\paragraph{Sentiment polarity (sentiment\_flip).} First-token accuracy is uninformative for \texttt{sentiment\_flip} (both polarities can begin with arbitrary tokens). We construct a sentiment contrast vector $\mathbf{d}_{\text{sent}} = (\bar{\mathbf{e}}_{\text{neg}} - \bar{\mathbf{e}}_{\text{pos}}) / \|\bar{\mathbf{e}}_{\text{neg}} - \bar{\mathbf{e}}_{\text{pos}}\|$, where $\bar{\mathbf{e}}_{\text{pos}}$ and $\bar{\mathbf{e}}_{\text{neg}}$ are mean token embeddings of the positive and negative first tokens in the sentiment-flip pairs. At each layer, we compute $\cos(\mathrm{LayerNorm}(\mathbf{h}_\ell), \mathbf{d}_{\text{sent}})$ for each example; polarity classification accuracy is the fraction with sign agreement to the expected polarity. We mark this metric explicitly in every figure and table caption that includes \texttt{sentiment\_flip} readability.

\paragraph{Multiple comparisons.} We apply Bonferroni correction wherever multiple hypothesis tests are conducted: within-task correlations ($k = 12$ tests, $\alpha_{\text{corrected}} = 0.05/12 = 0.0042$); logit-lens vs.\ FV paired tests ($k = 12$, $\alpha_{\text{corrected}} = 0.0042$); permutation tests for excess dissociation ($k = 12$ per model, same correction).

\paragraph{Permutation tests.} For each task on each model we test whether the observed dissociation rate exceeds chance using 1{,}000 permutations of the alignment--transfer pairing within the task. The $p$-value is the fraction of permutations yielding a dissociation rate at least as large as observed; per-task $p$-values are Bonferroni-corrected as above.

\paragraph{Tokenizer handling.} All three families (Llama BPE, Gemma SentencePiece, Mistral BPE) treat the leading-space variant of an output token differently. We encode every expected output token both with and without a leading space and use whichever produces a valid first token under each model's tokenizer. Worked examples of the encoding for each family: for \texttt{country\_capital} ``Paris,'' the Llama BPE tokenizer typically uses ``\textvisiblespace Paris,'' Gemma SentencePiece uses ``\textvisiblespace Paris,'' and Mistral BPE uses ``\textvisiblespace Paris''---all three resolving to a leading-space variant for words that follow non-trivial prefix text. For \texttt{first\_letter} ``A,'' we use the bare ``A'' token under all three.

\paragraph{Generation defaults.} Greedy decoding (no sampling), maximum 5 new tokens (10 for \texttt{sentiment\_flip}, 3 for \texttt{first\_letter}), end-of-sequence stopping enabled.

\paragraph{Precision and batching.} Model weights are bfloat16; FV mean accumulators use float32 to minimize numerical error. Batch sizes follow GPU profile (128 on H100, 256 on H200). To amortize the $\alpha$ sweep, the input batch is tiled across $\alpha$ values in a single forward pass (batch of $B$ prompts becomes a tiled batch of $B \cdot |\alpha\text{-strengths}|$ with per-strength multipliers applied at the residual-stream hook), producing all $\alpha$-strengths' generations in one pass. This is numerically identical to per-strength calls but 3--8$\times$ faster.

\FloatBarrier
\section{Compute Details}
\label{app:compute}

\begin{table}[!htbp]
\caption{Compute budget per model on NVIDIA H100 80GB. Activation patching is the dominant cost; the total per model is 15--24 hours.}
\label{tab:compute}
\centering
\small
\begin{tabular}{@{}llc@{}}
\toprule
\textbf{Stage} & \textbf{Estimated time} & \textbf{Hardware} \\
\midrule
Zero-shot + few-shot baselines & ${\sim}1$ h & GPU \\
FV extraction (1{,}536 FVs) & ${\sim}3$--$4$ h & GPU \\
Activation caching (zero-shot) & ${\sim}0.5$ h & GPU \\
Logit lens + FV vocab projection & ${\sim}0.3$ h & GPU \\
Steering evaluation (${\sim}98$K configurations) & ${\sim}10$--$18$ h & GPU \\
Tuned-lens training + evaluation & ${\sim}0.1$ h & GPU \\
\midrule
\textbf{Total per model} & ${\sim}\mathbf{15}$--$\mathbf{24}$ \textbf{h} & NVIDIA H100 80GB \\
\textbf{Total (6 models)} & ${\sim}\mathbf{4}$--$\mathbf{6}$ \textbf{days} & \\
\bottomrule
\end{tabular}
\end{table}

We additionally ran on NVIDIA H200 141GB for the largest steering sweeps; on H200 the steering-evaluation stage drops to ${\sim}6$--$10$ h per model, total ${\sim}10$--$15$ h. The full project (including preliminary experiments not reported in the main paper) consumed approximately 200 GPU-hours on H100 and 50 GPU-hours on H200.

\FloatBarrier
\section{Worked Example: country\_capital on Llama-3.1-8B Base}
\label{app:worked}

To make the methodology concrete, we walk through the full pipeline for a single (task, model) pair: \texttt{country\_capital} on Llama-3.1-8B Base.

\paragraph{Baseline accuracy.} Zero-shot accuracy: 0.08 (Llama Base only weakly retrieves capitals from bare prompts). Five-shot ICL: 0.84.

\paragraph{FV extraction.} For each of the 8 templates we extract per-layer FVs at all 32 residual-stream blocks. The IID-best layer averages L6.5 across templates, with templates T1 (``The capital of \{X\} is''), T4, T7, and T8 yielding strong FVs (per-template IID 0.84--0.88) and T2 (``\{X\}'s capital city is'') yielding the weakest (IID 0.22).

\paragraph{IID steering accuracy.} Mean across 8 templates at the best ($\alpha$, layer) per template: 0.665 (Table~\ref{tab:per_template_llama}). The FV redirects Llama Base's existing capability (zero-shot 0.08) to ICL-comparable accuracy.

\paragraph{Logit-lens decodability.} For zero-shot prompts of \texttt{country\_capital} on Llama Base, logit-lens top-10 accuracy is $<$0.01 at layers 1--20 and rises only weakly at late layers, peaking at L32 with mean top-10 = 0.056. The first-decodable layer (mean top-10 $\geq 0.10$) does not exist for this task on this model: the logit lens never crosses the threshold.

\paragraph{Steerability--decodability gap.} FV steering accuracy 0.880 (best template T4, L4) minus logit-lens best-layer top-10 0.056 = $-0.824$. This is the most dramatic steerability-without-decodability case in the entire 72-pair matrix.

\paragraph{Post-steering logit lens.} After applying the FV at L4, the maximum post-steering logit-lens top-10 across all subsequent layers and templates is 0.078---near zero. The FV steers the output without leaving any logit-lens-visible trace at intermediate layers, consistent with a non-representational (modulatory) mechanism on Llama Base.

\paragraph{Tuned-lens correction.} Per-layer diagonal translators are trained (mean improvement 11.4\%, max 16.5\%, $r$(depth, improvement) = $+0.898$). After applying the tuned lens, top-10 accuracy at the best layer is 0.056---unchanged. The dialect correction does not reveal hidden decodability.

\paragraph{Activation patching.} Country\_capital passes the IID gate, so patching is run. We sample 5 (source, target) template pairs, evaluate 50 IID examples per pair, and patch the residual stream at every layer. Recovery accuracy peaks at L12--L18 (max recovery 0.87 on T4$\to$T1), with mean across (case, layer) of 0.722. This places \texttt{country\_capital} in the ``mixed'' patching category (Table~\ref{tab:patching_full}): cross-template transfer is causally concentrated at specific layers in the recoverable range, but recovery is not perfect on every (source, target) pair.

\paragraph{Summary of the worked example.} Country\_capital on Llama Base is the cleanest illustration of our central claim: an FV that achieves 0.880 IID steering accuracy when applied at L4, redirects model behavior at specific layers (recoverable by patching), but is invisible to the model's own unembedding at every intermediate layer (logit lens 0.056 at peak), survives a dialect-correction robustness check (tuned lens 0.056), and produces near-zero post-steering deltas (0.078) on the logit-lens reading. The FV encodes a computational instruction that the model executes through pathways the unembedding cannot decode.

\end{document}